\documentclass[letterpaper]{article}
\usepackage{aaai24}  
\usepackage{times}  
\usepackage{helvet} 
\usepackage{courier} 
\usepackage[hyphens]{url} 
\usepackage{graphicx} 
\usepackage{natbib} 
\usepackage{caption}
\usepackage{mathtools}
\usepackage{graphicx,lipsum}
\usepackage{adjustbox}
\usepackage{multirow}

\usepackage{amssymb}
\usepackage{amsmath}

\usepackage{multicol}
\usepackage{subfig}

\usepackage{multirow}
\usepackage{adjustbox}

\usepackage{algorithm}
\usepackage[algo2e,ruled,vlined]{algorithm2e}

\usepackage{newfloat}
\usepackage{listings}
\DeclareCaptionStyle{ruled}{labelfont=normalfont,labelsep=colon,strut=off}
\floatstyle{ruled}
\newfloat{listing}{tb}{lst}{}
\floatname{listing}{Listing}

\title{Towards Diverse Perspective Learning \\ with Selection over Multiple Temporal Poolings}
\author{
    Jihyeon Seong\equalcontrib\textsuperscript{\rm 1},
    Jungmin Kim\equalcontrib\textsuperscript{\rm 1},
    Jaesik Choi\textsuperscript{\rm 1, \rm 2}
}
\affiliations{
    \textsuperscript{\rm 1}Korea Advanced Institute of Science and Technology (KAIST), South Korea \\
    \textsuperscript{\rm 2}INEEJI, South Korea\\
    \{jihyeon.seong, aldirl7, jaesik.choi\}@kaist.ac.kr
}

\usepackage{bibentry}

\begin{document}

\maketitle

\begin{abstract}
In Time Series Classification (TSC), temporal pooling methods that consider sequential information have been proposed. However, we found that each temporal pooling has a distinct mechanism, and can perform better or worse depending on time series data. We term this fixed pooling mechanism a single perspective of temporal poolings. In this paper, we propose a novel temporal pooling method with diverse perspective learning: Selection over Multiple Temporal Poolings (SoM-TP). SoM-TP dynamically selects the optimal temporal pooling among multiple methods for each data by attention. The dynamic pooling selection is motivated by the ensemble concept of Multiple Choice Learning (MCL), which selects the best among multiple outputs. The pooling selection by SoM-TP's attention enables a non-iterative pooling ensemble within a single classifier. Additionally, we define a perspective loss and Diverse Perspective Learning Network (DPLN). The loss works as a regularizer to reflect all the pooling perspectives from DPLN. Our perspective analysis using Layer-wise Relevance Propagation (LRP) reveals the limitation of a single perspective and ultimately demonstrates diverse perspective learning of SoM-TP. We also show that SoM-TP outperforms CNN models based on other temporal poolings and state-of-the-art models in TSC with extensive UCR/UEA repositories.
\end{abstract}

\section{Introduction}
Time Series Classification (TSC) is one of the most valuable tasks in data mining, and Convolutional Neural Network (CNN) with global pooling shows revolutionary success on TSC~\cite{LANGKVIST201411, ismail2019deep}. However, global pooling in TSC poses a significant challenge, as it disregards the fundamental characteristic of time series data, which is the temporal information, by compressing it into a single scalar value \citep{lecun1998gradient,yu2014mixed}. To tackle this issue, temporal pooling methods were introduced, which preserve the temporal nature of the time series at the pooling level \cite{lee2021learnable}.

Temporal pooling involves employing operations such as `maximum' (MAX) and `average' (AVG),  categorized by segmentation types: `no segment,' `uniform,' and `dynamic.' These segmentation types correspond respectively to Global-Temporal-Pooling (GTP), Static-Temporal-Pooling (STP), and Dynamic-Temporal-Pooling (DTP)~\cite{lee2021learnable}. We refer to each distinct pooling mechanism as a \textit{perspective} based on the segmentation types. However, we discovered that the most effective temporal pooling varies depending on the characteristics of the time series data, and there is no universally dominant pooling method for all datasets \cite{esling2012time}. This underlines the necessity for a learnable pooling approach adaptable to each data sample’s characteristics.

In this paper, we propose Selection over Multiple Temporal Poolings (SoM-TP). SoM-TP is a learnable ensemble pooling method that dynamically selects heterogeneous temporal poolings through an attention mechanism \cite{vaswani2017attention}. Aligned with our observation that a more suitable pooling exists for each data sample, a simple ensemble weakens the specified representation power \cite{lee2017confident}. Therefore, SoM-TP applies advanced ensemble learning, motivated by Multiple Choice Learning (MCL), that selects the best among the multiple pooling outputs ~\cite{guzman2012multiple}.

MCL is a selection ensemble that generates $M$ predictions from multiple instances, computes the oracle loss for the most accurate prediction, and optimizes only the best classifier. Capitalizing on the advantage of deep networks having access to intermediate features, SoM-TP ensembles diverse pooling features in a single classifier. To achieve non-iterative optimization, SoM-TP dynamically selects the most suitable pooling method for each data sample through attention, which is optimized by \textit{Diverse Perspective Learning Network (DPLN)} and \textit{perspective loss}. DPLN is a sub-network that utilizes all pooling outputs, and the perspective loss reflects DPLN’s result to make a regularization effect. Finally, the CNN model based on SoM-TP forms fine representations through diverse pooling selection, allowing it to capture both the `global' and `local' features of the dataset.
\begin{figure}
    \centering
    \subfloat[GTP]{\includegraphics[width=0.13\textwidth, height=0.17\textwidth]{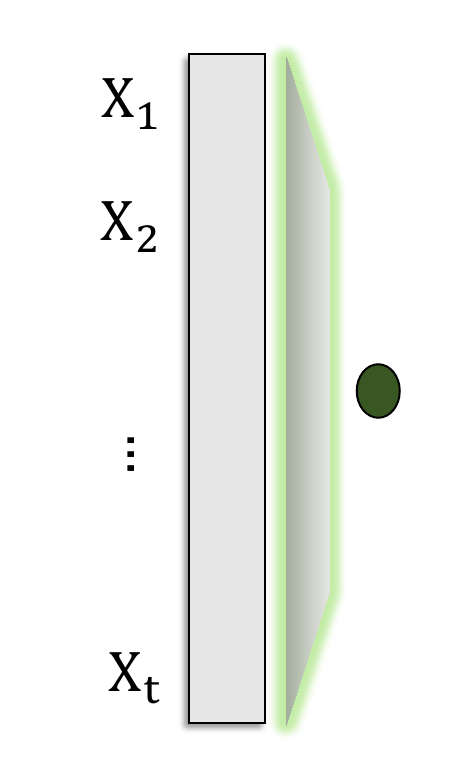}}
    \subfloat[STP]{\includegraphics[width=0.13\textwidth, height=0.17\textwidth]{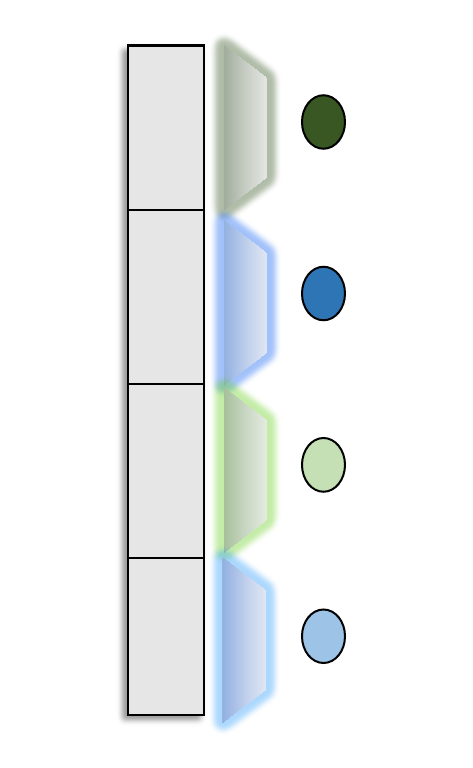}}
    \subfloat[DTP]{\includegraphics[width=0.205\textwidth, height=0.18\textwidth]{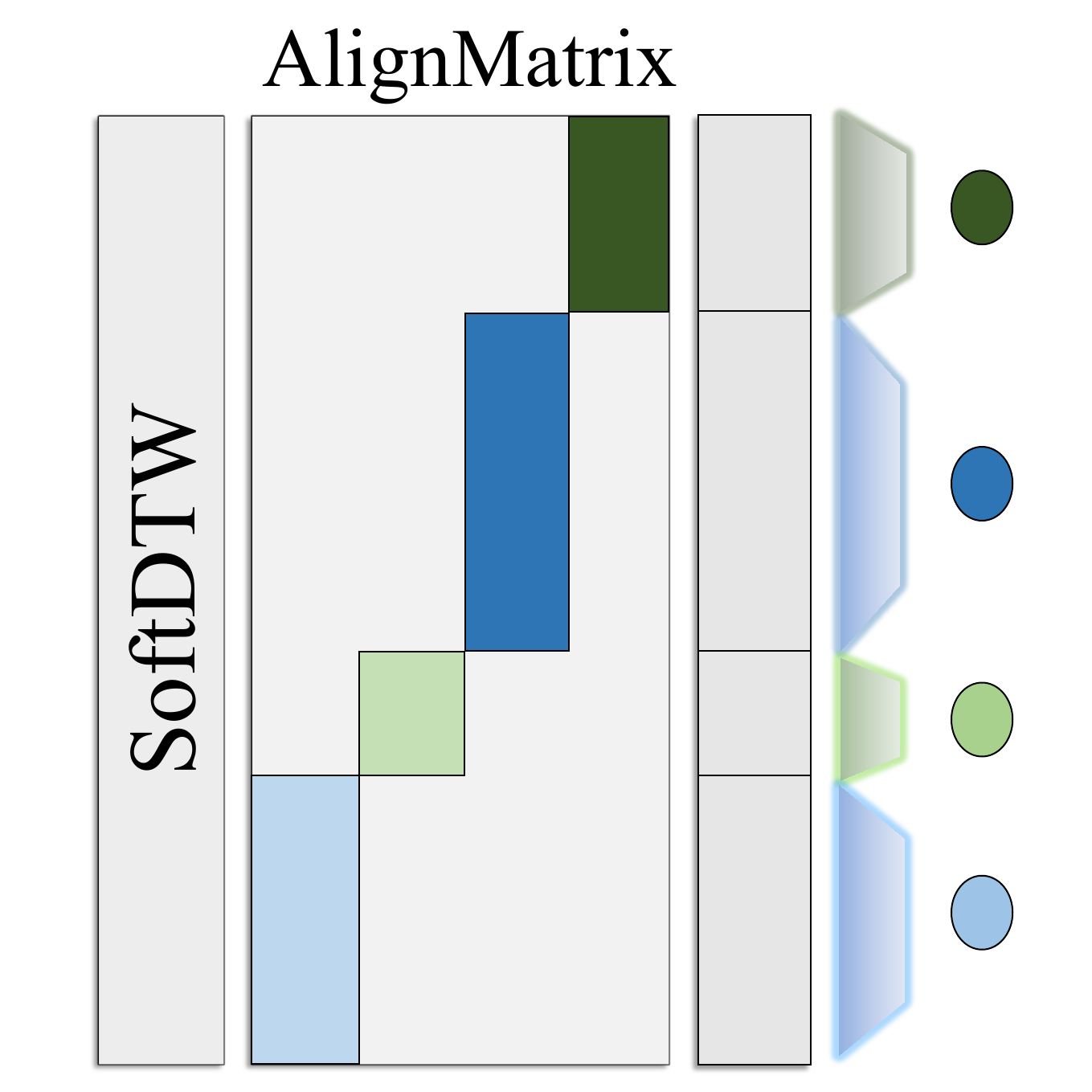}}
    \caption{Perspectives of Temporal Poolings. Depending on segmentation types, each temporal pooling generates different pooling outputs and has different \textit{perspectives}.}
    \label{fig:mesh1}
\end{figure}

Recognizing the crucial role of pooling in selecting the most representative values from encoded features in CNNs, we have chosen CNNs as the suitable model for our study. We apply our new selection ensemble pooling to Fully Convolutional Networks (FCNs) and Residual Networks (ResNet), which show competitive performances in TSC as a CNN-based model \cite{wang2017time,ismail2019deep}. SoM-TP outperforms the existing temporal pooling methods and state-of-the-art models of TSC both in univariate and multivariate time series datasets from massive UCR/UEA repositories. We also provide a detailed analysis of the diverse perspective learning result by Layer-wise Relevance Propagation (LRP) \cite{bach2015pixel} and the dynamic selection process of SoM-TP. To the best of our knowledge, this is the first novel approach to pooling-level ensemble study in TSC.

Therefore, our contributions are as follows:

\begin{itemize}
\item We investigate data dependency arising from distinct perspectives of existing temporal poolings.
\item We propose SoM-TP, a new temporal pooling method that fully utilizes the diverse temporal pooling mechanisms through an MCL-inspired selection ensemble.
\item We employ an attention mechanism to enable a non-iterative ensemble in a single classifier.
\item We define DPLN and perspective loss as a regularizer to promote diverse pooling selection.
\end{itemize}

\section{Background}
\label{sec2}

\subsection{Different Perspectives between Temporal Poolings}
\label{subsec2.1}
\subsubsection{Convolutional Neural Network in TSC}
In TSC, CNN outperforms conventional methods, such as nearest neighbor classifiers~\cite{Yuan2019} or COTE~\cite{7069254,7837946}, by capturing local patterns of time series ~\cite{ismail2019deep}. 

The TSC problem is generally formulated as follows: a time series data $\mathbf{T} = \{(\mathbf{X}_1, y_1), ..., (\mathbf{X}_t, y_t)\}$, where $\mathbf{X} \in \mathbb{R}^\mathbf{d\times t}$ of length $\mathbf{t}$ with $\mathbf{d}$ variables and $y\in \{1, ..., C\}$ from $C$ classes. Then the convolution stack $\Phi$ of out channel dimension $\mathbf{k}$ encodes features as hidden representations with temporal position information $\mathbf{H} = \{h_0, ..., h_t\} \in \mathbb{R}^{\mathbf{k} \times \mathbf{t}}$~\cite{lee2021learnable, wang2017time}.
\begin{equation}
\mathbf{H} = \Phi(\mathbf{T})
\end{equation}

After convolutional layers, global pooling plays a key role with two primary purposes: 1) reducing the number of parameters for computational efficiency and preventing overfitting, and 2) learning position invariance. For this purpose, pooling combines the high-dimensional feature outputs into low-dimensional representations \cite{gholamalinezhad2020pooling}. However, global pooling presents an issue of losing temporal information, which has led to the development of temporal pooling methods \cite{lee2021learnable}. We investigate different mechanisms of temporal poolings, which we refer to as \textit{perspective}.

\subsubsection{Global Temporal Pooling}
GTP pools only one representation $\mathbf{p_g} = [p_1] \in \mathbb{R}^{\mathbf{k}\times\mathbf{1}}$ in the entire time range. GTP ignores temporal information by aggregating the $\mathbf{H}$ to $\mathbf{p}_g = h$: the global view.
\begin{equation}
\mathbf{p}_g = pool_g(\mathbf{H})
\end{equation}
GTP effectively captures globally dominant features, such as trends or the highest peak, but has difficulty capturing multiple points dispersed on a time axis. To solve this constraint, temporal poolings based on sequential segmentation have been proposed: STP and DTP \cite{lee2021learnable}. Both have multiple local segments with the given number $\mathbf{n} \in \mathbb{Z}^+$: the local view.

\subsubsection{Static Temporal Pooling}
STP divides the time axis equally into $\mathbf{n}$ segments with a length $\ell = \frac{t}{n}$, where \\ $\mathbf{\bar{H} = \{ \mathbf{h}_{0:\ell}, \mathbf{h}_{\ell:2\ell}, ... , \mathbf{h}_{(n-1)\ell:n\ell} \}}$ and $\mathbf{p_s} = [p_1, ..., p_n] \in \mathbb{R}^{\mathbf{k} \times \mathbf{n}}$. Note that $\mathbf{h}_\ell$ retains temporal information, but there is no consideration of the temporal relationship between time series in the segmentation process: the uniform local view.
\begin{equation}
\mathbf{p}_s = pool_s(\mathbf{\bar{H}})
\end{equation}
STP functions well on a recursive pattern, such as a stationary process. However, forced uniform segmentation can divide important consecutive patterns or create unimportant segmentations. This inefficiency causes representation power to be distributed to non-informative regions.

\subsubsection{Dynamic Temporal Pooling}
DTP is a learnable pooling layer optimized by soft-DTW \cite{cuturi2017soft} for dynamic segmentation considering the temporal relationship. By using the soft-DTW layer, $\mathbf{H}$ is segmented in diverse time lengths $\bar{\ell} = [\ell_1, \ell_2, ...,\ell_n]$, where $\mathbf{t} = \sum \bar{\ell}$. Finally, the optimal pooled vectors $\mathbf{p}_d = [p_{\ell_1}, ..., p_{\ell_n}] \in \mathbb{R}^{\mathbf{k} \times \mathbf{n}}$ are extracted from each segment of $\Bar{\mathbf{H}}_{\bar{\ell}}$, where $\Bar{\mathbf{H}}_{\Bar{\ell}} = \{\mathbf{h}_{\ell_1}, \mathbf{h}_{\ell_2}, ..., \mathbf{h}_{\ell_n}\}$; the dynamic local view.
\begin{equation}
\mathbf{p}_d = pool_d({\Bar{\mathbf{H}}_{\Bar{\ell}}})
\end{equation}
DTP has the highest complexity in finding different optimal segmentation lengths, enabling the pooling to fully represent segmentation power. However, since DTP is based on temporally aligned similarity of hidden features with a constraint that a single time point should not be aligned with multiple consecutive segments, the segmentation can easily divide informative change points that need to be preserved in time series patterns (Appendix. DTP Algorithm).

\begin{figure*}
    \centering
    \includegraphics[width=0.9\textwidth]{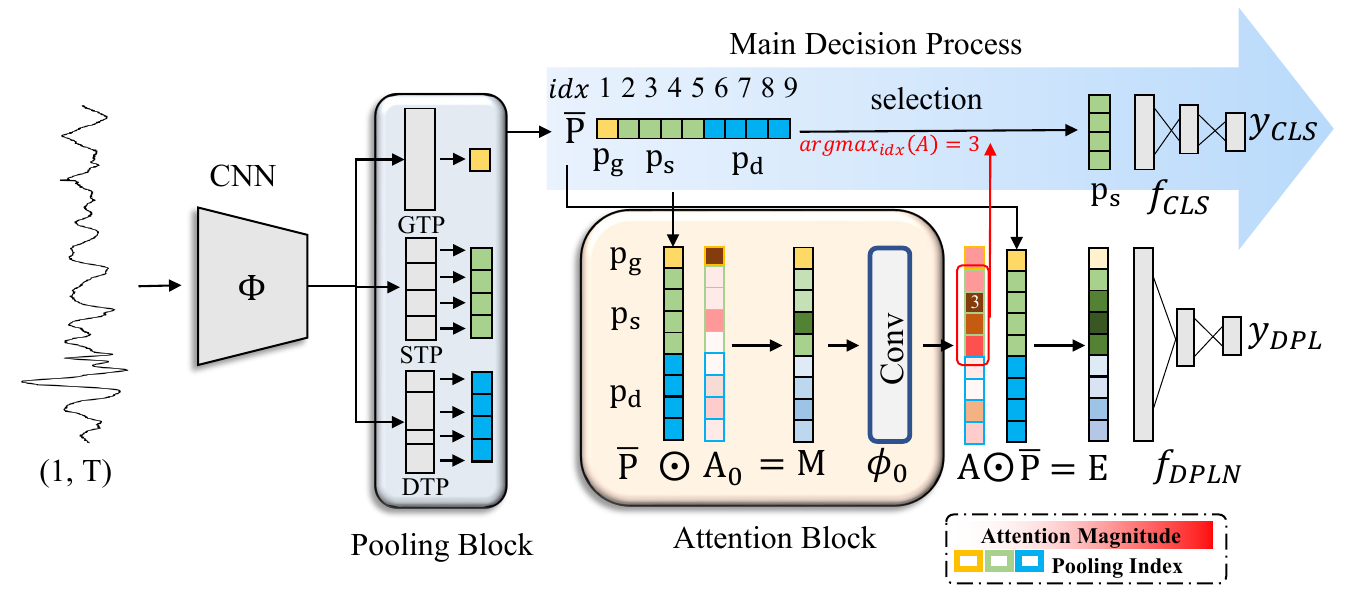}
    \caption{SoM-TP Architecture. Diverse Perspective Learning based on selection-ensemble is achieved as follows: The aggregated output of all pooling, $\bar{\mathbf{P}}$ is passed to the attention block to calculate the attention score $\mathbf{A}$. In the attention block, a weighted pooling output $\mathbf{M}$ is formed by the multiplication of $\bar{\mathbf{P}}$ and a learnable weight vector $\mathbf{A_0}$. After $\mathbf{M}$ passes through the convolutional layer $\phi_0$, the attention score $\mathbf{A}$ is drawn out as an encoded weight vector. Using the index of the highest attention score (here, index 3), pooling for the CLS network is selected. Next, the parameters are updated with the following procedure: 1) DPLN uses the ensembled vector $\mathbf{E}$, whereas CLS network uses only the selected pooling output (here, $\mathbf{p_s}$); 2) Each network predicts $y_{CLS}$ and $y_{DPL}$ respectively, and $y_{DPL}$ is used in the perspective loss to work as a regularizer; 3) With these two outputs, the model is optimized with diverse perspectives while selecting the proper pooling method for each batch.}
    \label{fig:mesh2}
\end{figure*}

\subsubsection{Limitation of Single Perspective}
Traditional temporal pooling methods only focus on a single perspective when dealing with hidden features $\mathbf{H}$. A global perspective cannot effectively capture multiple classification points, while a local perspective struggles to emphasize a dominant classification point. Consequently, datasets that require the simultaneous capture of dominant and hidden local features from diverse viewpoints inevitably exhibit lower performance when using a single perspective. Motivated by these limitations, we propose a novel pooling approach that fully leverages diverse perspectives.

\subsection{Multiple Choice Learning for Deep Temporal Pooling}
\label{sec2.2}

The traditional ML-based ensemble method focuses on aggregating multiple outputs. However, the aggregation of the simple ensemble makes outputs smoother, due to the generalization effect~\cite{lee2017confident}. To solve this limitation, MCL has been proposed as an advanced ensemble method that selects the best among multiple outputs using oracle loss~\cite{guzman2012multiple}. More formally, MCL generates $M$ solutions $\hat{Y}_i = (\hat{y}_i^1, ..., \hat{y}_i^M)$, and learns a mapping $g: \mathcal{X} \rightarrow \mathcal{Y}^M$ that minimizes oracle loss $min_m \ell(y_i, \hat{y}_i^m)$. The ensemble mapping function $g$ consists of multiple predictors, $g(x) = \{f_1(x), f_2(x), ..., f_M(x)\}$~\cite{lee2016stochastic}.

The effects of diverse solution sets in MCL can be summarized as addressing situations of `Ambiguous evidence’ and `Bias towards the mode’. `Ambiguous evidence’ refers to situations with insufficient information to make a definitive prediction. In such cases, presenting a small set of reasonable possibilities can alleviate the over-confidence problem of deep learning \cite{7298640}, rather than striving for a single accurate answer. The other situation is `Bias towards the mode’, indicating the model’s tendency to learn a mode-seeking behavior to reduce the expected loss across the entire dataset. When only a single prediction exists, the model eventually learns to minimize the average error. In contrast, MCL generates multiple predictions, allowing some classifiers to cover the lower-density regions of the solution space without sacrificing performance on the high-density regions \cite{lee2016stochastic}.

MCL faces computational challenges in deep networks due to the iterative optimization process of the oracle loss, with a complexity of $\mathcal{O}(N^2)$. Although sMCL has partially alleviated this issue through stochastic gradient descent, the method still requires identifying the best output among multiple possibilities \cite{lee2016stochastic}. CMCL, which is another approach to address MCL's over-confidence problem, cannot be applied at the feature level due to optimization at the output level \cite{lee2017confident}. In summary, the integration of MCL into the pooling level is not feasible due to the structural constraints imposed by the oracle loss design. To overcome this challenge, we establish a model structure to incorporate the concept of MCL into a pooling-level ensemble.

\section{Selection over Multiple Temporal Pooling}
\label{sec3}
\subsection{SoM-TP Architecture and Selection Ensemble} 
\label{sec3.1}
Diverse Perspective Learning (DPL) is achieved by dynamically selecting heterogeneous multiple temporal poolings in a single classifier. The overview architecture, as illustrated in Figure \ref{fig:mesh2}, consists of four parts: 1) a common feature extractor with CNN $\Phi$; 2) a pooling block with multiple temporal pooling layers; 3) an attention block with an attention weight vector $\mathbf{A}_0$, and a convolutional layer $\phi_0$, as well as 4) Fully Connected layers (FC): a classification network (CLS) $f_{CLS}$ and a \textit{DPLN} $f_{DPLN}$. Through these modules, SoM-TP can cover the high probability prediction space, which is aligned with MCL where multiple classifiers are trained to distinguish specific distributions \cite{lee2016stochastic}.

\newpage
\subsubsection{DPL Attention}
SoM-TP ensembles multiple temporal pooling within a single classifier. The advantage of using a single classifier lies in the absence of the need to compare prediction outputs, making it non-iterative and computationally efficient, in contrast to MCL. Through the attention mechanism \cite{vaswani2017attention}, we achieve a `comparison-free' ensemble by dynamically selecting the most suitable temporal pooling for each batch.

DPL attention is an extended attention mechanism that simultaneously considers two factors: the overall dataset and each data sample. In Figure \ref{fig:mesh2}, the attention weight vector $\mathbf{A}_0$ is learned in the direction of adding weight to specific pooling, which has a minimum loss for every batch. Consequently, $\mathbf{A}_0$ has a weight that reflects the entire dataset.  

Next, a convolutional layer $\phi_0$ is used to reflect a more specific data level. As shown in Algorithm \ref{main algo}, the weighted pooling vector $\mathbf{M}$ which is the element-wise multiplication between $\mathbf{A}_0$ and $\Bar{\mathbf{P}}$ is given as an input to $\phi_0$. By encoding $\mathbf{M}$ through $\phi_0$, $\mathbf{A}$ can reflect more of each batch characteristic, not just following the dominant pooling in terms of the entire dataset. As a result, this optimized pooling selection by attention can solve the `Bias towards the mode' problem by assigning the most suitable pooling for each data batch, which is aligned with learning the multiple experts in MCL.

\subsection{Diverse Perspective Learning Network and Perspective Loss} 
\label{sec3.2}

To optimize DPL attention, we introduce DPLN and perspective loss. DPLN is a sub-network that utilizes the weighted aggregation ensemble $\mathbf{E}$. The main role of DPLN is regularization through perspective loss. In contrast, the CLS network, which predicts the main decision, does not directly utilize attention $\mathbf{A}$. Instead, it employs a chosen pooling feature output (denoted as $\mathbf{p_s}$ in Figure \ref{fig:mesh2}), determined by the index with the biggest score in $\mathbf{A}$.

\subsubsection{Perspective Loss}
Perspective loss serves as a cost function to maximize the utilization of the sub-network, DPLN, through network tying between the two FC networks. Ultimately, it aims to prevent the CLS network from converging to one dominant pooling and to maintain the benefits of ensemble learning continuously. 

To achieve its purpose, perspective loss is designed as the sum of DPLN cross-entropy loss and the Kullback-Leibler (KL) divergence between $y_{CLS}$ and $y_{DPL}$. Specifically, KL divergence works similarly to CMCL’s KL term, which regulates one model to be over-confident through a uniform distribution, while the KL term of perspective loss regulates based on DPLN \cite{lee2017confident}.

\begin{equation}
\begin{split}
&KL(y_{CLS}, y_{DPL}) = y_{DPL} \cdot log \frac{y_{DPL}}{y_{CLS}}, \\
&\mathcal{L}_{DPLN}(\{\mathcal{W}_{\Phi}\}, \{\mathbf{W}^{(dpln)}\}) = -\frac{1}{t} \sum_{n=1}^t log P(y=y_n | \mathbf{X}_n), \\
&\mathcal{L}_{perspective} = KL(y_{CLS}, y_{DPL}) + \mathcal{L}_{DPLN}, \\
\end{split}
\end{equation}

\begin{algorithm}[H]
\caption{SoM-TP selecting algorithm}
\label{main algo}    
\SetAlgoLined
\DontPrintSemicolon

  \SetKwFunction{FMain}{Attention Block}
  \SetKwFunction{pooling}{Pooling Block}
 
  \SetKwProg{Fn}{Function}{:}{}
  \Fn{\FMain{$\mathbf{H}$}}{

        $\triangleright$ select proper temporal pooling by attention $\mathbf{A} \in \mathbb{R}^{1\times 3\mathbf{n}}$ 

        $\triangleright$ attention weight $\mathbf{A}_0 \in \mathbb{R}^{1\times 3\mathbf{n}}$ is initialized as zero 

        $\triangleright$ GTP, STP, DTP: $pool_g, pool_s, pool_d$ 

        $\triangleright$ convolutional encoding layer: $\phi_0$ 

        \Fn{\pooling{$\mathbf{H}$}}{
            $\triangleright$ convolutional hidden feature:  $\mathbf{H} \in \mathbb{R}^{k \times t}$ 

            $\triangleright$ static segmented hidden feature: \\ $\mathbf{\bar{H} = \{ \mathbf{h}_{0:\ell}, \mathbf{h}_{\ell:2\ell}, ... , \mathbf{h}_{(n-1)\ell:n\ell} \}}, \ell = \frac{t}{n}$ 

            $\triangleright$ dynamic segmented hidden feature: \\ $\Bar{\mathbf{H}}_{\Bar{\ell}} = \{\mathbf{h}_{\ell_1}, \mathbf{h}_{\ell_2}, ..., \mathbf{h}_{\ell_n}\}$, \\
            $\bar{\ell} = [\ell_1, \ell_2, ...,\ell_n]$, where $\mathbf{t} = \sum \bar{\ell}$ 

            $\triangleright$ pooling outputs: \\
            $\mathbf{p}_g, \mathbf{p}_s, \mathbf{p}_d = pool_g(\mathbf{H}), pool_s(\Bar{\mathbf{H}}), pool_d(\Bar{\mathbf{H}}_{\Bar{\ell}})$ 
            
            \textbf{return} $\mathbf{p}_g, \mathbf{p}_s, \mathbf{p}_d$ 
        }
        
        $\Bar{\mathbf{P}} = [\mathbf{p}_g, \mathbf{p}_s, \mathbf{p}_d]$ 

        $\mathbf{M} = \mathbf{A}_0 \odot \Bar{\mathbf{P}}$ 
        
        $\mathbf{A} = \phi_0(\mathbf{M})$, where $x \in \mathbf{A}$ 
        \[idx =\begin{cases}
                argmax_i(y), \text{ where } y_i = \frac{exp(x_i)}{\sum_i exp(x_i)} \\ 
                argmax_j(y), \text{ where } y_j = \frac{\sum_j^\mathbf{n} x_{(j|n)}}{\mathbf{n}}
                \end{cases} \]

        \textbf{return} $\mathbf{p} = \Bar{\mathbf{P}}(idx), \mathbf{E} = \mathbf{A} \odot \Bar{\mathbf{P}}$ 
    }
\end{algorithm}

where input time series $\{(\mathbf{X}_1, y_1), ..., (\mathbf{X_t}, y_t)\}$, $\Phi$ with learnable parameter $\mathcal{W}_{\Phi}$ of CNN, $y_{CLS} \in \mathbb{R}^{1\times c}$ from the `CLS network' $\mathbf{W}^{(cls)}$, and $y_{DPL} \in \mathbb{R}^{1\times c}$ from the DPLN $\mathbf{W}^{(dpln)}$. Then, we set first $f_{DPLN}$ weight matrix $\mathbf{W}_0^{(dpln)} = [\mathbf{w}_1^{(\mathbf{p}_g)}, ... , \mathbf{w}_{2\mathbf{n}}^{(\mathbf{p}_s)}, ..., \mathbf{w}_{3\mathbf{n}}^{(\mathbf{p}_d)}] \in \mathbb{R}^{k \times 3\cdot \mathbf{n}}$, where $\mathbf{w}^{(\mathbf{p})} \in \mathbb{R}^k$ is weight matrix of each latent dimension $k$ of pooling $\mathbf{p}_i$, whereas $\mathbf{W}_0^{(cls)} = [\mathbf{w}_1^{(c)}, ..., \mathbf{w}_{\mathbf{n}}^{(c)}] \in \mathbb{R}^{k \times \mathbf{n}}$ is the first $f_{CLS}$ weight matrix~\cite{lee2021learnable}. Note that the results of GTP are repeated $\mathbf{n}$ times to give the same proportion for each pooling by attention weight.

Therefore, the final loss function of the SoM-TP is designed as follows,

\begin{equation}
\begin{split}
&\mathcal{L}_{CLS}(\{\mathcal{W}_{\Phi}\}, \{\mathbf{W}^{(cls)}\}) = -\frac{1}{t} \sum_{n=1}^t log P(y=y_n | \mathbf{X}_n), \\
&\mathcal{L}_{cost}(\{\mathcal{W}_{\Phi}\}, \{\mathbf{W}\}) = \mathcal{L}_{CLS} + \lambda \cdot \mathcal{L}_{perspective},
\end{split}
\end{equation}

where $\{\mathbf{W}^{(cls)}, \mathbf{W}^{(dpln)}, \mathbf{A}_0, \phi_0\} \in \mathbf{W}$ are learnable parameters. Prioritizing classification accuracy, the loss $\mathcal{L}_{CLS}$ is computed, and $\mathcal{L}_{perspective}$ is added with $\lambda$ decay. 

\begin{table*}[t]
\centering
    \begin{adjustbox}{width=0.93\textwidth}
    \begin{tabular}{lll|lllll|lllll}
        \noalign{\smallskip}\noalign{\smallskip}
        \bf CNN & \multicolumn{2}{l}{\bf POOL (type)} & \multicolumn{5}{l}{\bf UCR (uni-variate)} & \multicolumn{5}{l}{\bf UEA (multi-variate)} \\
        & \multicolumn{2}{l|}{\bf MAX} & ACC & F1 macro & ROC AUC & PR AUC & Rank & ACC & F1 macro & ROC AUC & PR AUC & Rank\\
        \hline
        \multirow{5}{*}{FCN} & \multicolumn{2}{l|}{GTP} & 0.6992 & 0.6666 & 0.8662 & 0.7406 & 3.6 & 0.6558 & 0.6213 & 0.7841 & 0.6854 & 3.9\\
                 & \multicolumn{2}{l|}{STP} & \underline{0.7462} & \underline{0.7133} & \underline{0.8924} & \underline{0.7889} & \underline{2.7} &  \underline{0.6801} & \underline{0.6603} & 0.8001 & 0.6984 & \underline{3.0}\\
                 & \multirow{2}{*}{DTP} & euc & 0.7406 & 0.7123 & 0.8897 & 0.7782 & 3.1 & 0.6795 & 0.6559 & \underline{0.8129} & \underline{0.7163} & 3.4\\
                 & & cos & 0.7335 & 0.7062 & 0.8879 & 0.7768 & 2.9 & 0.6702 & 0.6314 & 0.8061 & 0.7022 & 3.1\\
                \cline{2-13}
                 & \multicolumn{2}{l|}{SoM-TP} & \bf 0.7556 & \bf 0.7241 & \bf 0.9026 & \bf 0.8000 & \bf 2.6 & \bf 0.6920 & \bf 0.6621 & 0.8105 & 0.7099 & \bf 2.4\\
        \hline
        \multirow{5}{*}{ResNet} & \multicolumn{2}{l|}{GTP} & 0.7227 & 0.6952 & 0.8837 & 0.7654 & 3.5 & 0.6423 & 0.6083 & 0.7798 & 0.6769 & 3.5\\
                 & \multicolumn{2}{l|}{STP} & 0.7420 & 0.7126 & 0.8880 & 0.7864 & \underline{2.9} & \underline{0.6717} & \underline{0.6383} & 0.7962 & 0.6934 & \underline{2.8}\\
                 & \multirow{2}{*}{DTP} & euc & \underline{0.7456} & 0.7197 & 0.8939 & \underline{0.7846} & 3.0 & 0.6567 & 0.6271 & \underline{0.7981} & \underline{0.6968} & 3.1\\
                 & & cos & 0.7452 & \underline{0.7198} & \underline{0.8945} & 0.7829 & 3.0 & 0.6534 & 0.6377 & 0.7895 & 0.6832 & 3.0\\
                 \cline{2-13}
                 & \multicolumn{2}{l|}{SoM-TP} & \bf 0.7773 & \bf 0.7489 & \bf 0.9182 & \bf 0.8261 & \bf 2.4 & \bf 0.6769 & \bf 0.6387 & \bf 0.8016 & \bf 0.7033 & \bf 2.5\\
        \hline
        \multicolumn{8}{l}{\small *This table is for SoM-TP pooling selection operation type MAX.}
    \end{tabular}
    \end{adjustbox}
    \caption{SoM-TP Comparison with Single Perspective Temporal Poolings. The table presents the effectiveness of the selection ensemble of SoM-TP compared to traditional temporal poolings on operation type MAX. The best performances where SoM-TP outperforms others are bolded, and the best performances of other temporal poolings are underlined.}
    \label{main:t2}
\end{table*}

As a result, SoM-TP can address the `Ambiguous evidence' problem through DPLN and perspective loss. In a pooling ensemble, the `Ambiguous evidence' can be conceived as a scenario where a single pooling is not dominant. Even though SoM-TP selects only one pooling, $y_{DPL}$ in the perspective loss enables the model to consider the importance of other poolings.

\subsection{Optimization}
\label{sec3.3}
For attention weight $\mathbf{A}_0$, SoM-TP proceeds with additional optimization: a dot product similarity to regulate $\mathbf{A}_0$. The similarity term is defined as, 

\begin{equation}
\begin{split}
&\mathcal{L}_{attn} = - y_{CLS} \cdot y_{DPL},
\end{split}
\end{equation}
Due to the KL-divergence cost function in perspective loss, the CLS network and DPLN can be overly similar during the optimization process. As the additional regulation for output over-similarity, $\mathcal{L}_{attn}$ plays the opposite regulation to the perspective loss. Note that the dot-product similarity considers both the magnitude and direction of two output vectors.

Finally, the overall optimization process is as follows,
\begin{equation}
\begin{split}
&\mathbf{A_0} \leftarrow \mathbf{A_0} - \eta \cdot \partial \mathcal{L}_{attn} / \partial \mathbf{A_0}, \\
&\mathcal{W}_{\Phi} \leftarrow \mathcal{W}_{\Phi} - \eta \cdot \partial \mathcal{L}_{cost} / \partial \mathcal{W}_{\Phi}, \\
&\mathbf{W} \leftarrow \mathbf{W} - \eta \cdot \partial \mathcal{L}_{cost} / \partial \mathbf{W}. \\
\end{split}
\end{equation}

As a result, even with a non-iterative optimization process, SoM-TP learns various perspectives through DPL attention, DPLN, and perspective loss. Consequently, $\Phi$ reflects $f_{CLS}$ and $f_{DPLN}$ relatively while minimizing the similarity between each network output.

\begin{figure*}
    \centering
    \subfloat[GTP most selected]{\includegraphics[width=0.33\textwidth]{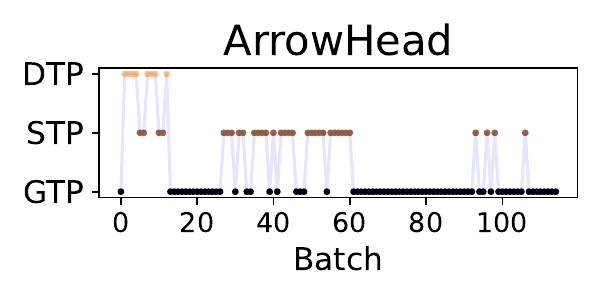}}
    \subfloat[STP most selected]{\includegraphics[width=0.33\textwidth]{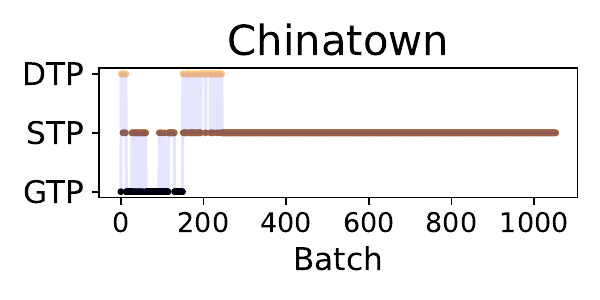}}
    \subfloat[DTP most selected]{\includegraphics[width=0.33\textwidth]{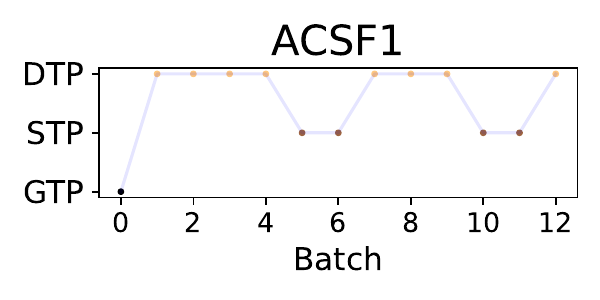}}
    \caption{Dynamic Pooling Selection in SoM-TP. This figure represents the graph of dynamic selection in the FCN SoM-TP MAX on the UCR repository: ArrowHead, Chinatown, and ACSF1.}
    \label{fig:3}
\end{figure*}

\begin{table*}[t]
    \centering
    \resizebox{0.99\textwidth}{!}{
    \begin{tabular}{c|cccc|cccc}
        \hline
         \multirow{2}{*}{Methods} & \multicolumn{4}{c|}{UCR} & \multicolumn{4}{c}{UEA}\\
         & Baseline & SoM-TP wins & Tie & Rank & Baseline & SoM-TP wins & Tie & Rank  \\
         \hline
         Vanilla-Transformer \cite{vaswani2017attention} & 10 & \textbf{99} & 4 &  5.2 & 3 & \textbf{18} & 1 & 4.4\\
         TCN \cite{bai2018empirical} & 28 & \textbf{80} & 5 & 4.0 & 4 & \textbf{17} & 1 & 4.3\\
         TST \cite{zerveas2021transformer} & 32 & \textbf{78} & 3 & 3.8 & 6 & \textbf{15} & 1 & 3.6\\
         ConvTran \cite{ConvTran2023} & 35 & \textbf{71} & 7 & 3.1 & 8 & \textbf{13} & 1 & 3.1\\
         MLSTM-FCN \cite{karim2019multivariate} & 46 & \textbf{60} & 6 & 2.6 & 6 & \bf 12 & 4 & 3.2\\
         \hline
         SoM-TP - MAX & - & - & - & \textbf{2.5} & - & - & - & \textbf{2.5}\\
         \hline
         \multicolumn{2}{c}{}
    \end{tabular}}
    \bigskip
    \resizebox{0.99\textwidth}{!}{
    \begin{tabular}{c|cccc|cccc}
        \hline
         \multirow{2}{*}{Methods} & \multicolumn{4}{c|}{UCR} & \multicolumn{4}{c}{UEA}\\ 
         & Acc & F1-score & ROC AUC & PR AUC & Acc & F1-score & ROC AUC & PR AUC \\
         \hline
         ROCKET \cite{dempster2020rocket} & \underline{0.7718} & \underline{0.7478} & 0.8899 & 0.7841 & {0.6785} & \underline{0.6592} & 0.7926 & 0.6940 \\ 
         InceptionTime \cite{ismail2020inceptiontime} & {0.7713} & {0.7455} & \underline{0.9056} & \underline{0.8164} & 0.6612  & 0.6360 & {0.7984} & \underline{0.7106}\\
         OS-CNN \cite{tang2020omni} & 0.7663 & 0.7324 & 0.9005 & {0.8139} & \underline{0.6808} & {0.6547} & \textbf{0.8118} & \textbf{0.7137} \\
         DSN \cite{xiao2022dynamic} & 0.7488 & 0.7230 & 0.8838 & 0.7968 & 0.5648 & 0.5433 & 0.7575 & 0.6265\\
         \hline
         SoM-TP - MAX & \bf 0.7773 & \bf 0.7489 & \bf 0.9182 & \bf 0.8261 & \bf 0.6920 & \bf 0.6621 & \underline{0.8105} & 0.7099 \\
         \hline
    \end{tabular}}
    \caption{SoM-TP Comparison with Advanced TSC Methods. This table compares the performance of SoM-TP with advanced TSC models that leverage temporal information and those that exploit scale-invariant properties, respectively. The best performances, where SoM-TP beat others, are bolded, and the best performances among other models are underlined.}
    \label{tab:2}
\end{table*}

\begin{table}[h]
    \centering
    \begin{adjustbox}{width=0.49\textwidth}
    \begin{tabular}{c|cccc|ccc|c}
    \hline
    \multirow{2}{*}{Type} & \multicolumn{4}{c|}{SoM-TP Modules} & \multicolumn{3}{c|}{Rank}  & \multirow{2}{*}{Acc} \\
    & $\mathbf{A}_0$ & $\phi_0$ & DPLN & $\mathcal{L}_{attn}$ & 1 & 2 & 3 &  \\
    \hline
         only $\phi_0$ & & \checkmark & & & 8 & 15 & 17 & 0.6966 \\
         only $\mathbf{A}_0$ & \checkmark & & & & 4 & 9 & 24 & 0.6963 \\
         DPL Attention& \checkmark & \checkmark & & & 6 & 7 & 23 & 0.6974\\
         \hline
         DPLN w/o $\mathbf{A}_0$ & & \checkmark & \checkmark &  & 6 & 14 & 29 & 0.7047 \\
         DPLN &\checkmark & \checkmark & \checkmark & & 23 & 27 & 26 & 0.7399 \\
        \hline
        SoM-TP &\checkmark & \checkmark & \checkmark & \checkmark & \bf 42 & 20 & 6 & \bf 0.7503 \\
    \hline
    \end{tabular}
    \end{adjustbox}
    \caption{SoM-TP Module Ablation Study.}
    \label{tab:small}
\end{table}

\section{Experiments}
\label{sec4}
\subsection{Experimental Settings}
\label{sec4.1}
For the extensive evaluation, 112 univariate and 22 multivariate time series datasets from the UCR/UEA repositories are used~\cite{bagnall2018uea,dau2019ucr}; collected from a wide range of domains and publicly available. To ensure the validity of our experiments, we exclude a few datasets from the UCR/UEA repositories due to the irregular data lengths. While zero padding could resolve this, it might cause bias in some time series models.

All temporal pooling methods have the same CNN architecture. FCN and ResNet are specifically designed as a feature extractor~\cite{wang2017time}, and temporal poolings are constructed with the same settings: normalization with BatchNorm~\cite{batchnorm}, activation function with ReLU, and optimizer with Adam~\cite{adam}. The validation set is made from $20\%$ of the training set for a more accurate evaluation. In the case of imbalanced classes, a weighted loss is employed. The prototype number $\mathbf{n}$ is searched in a greedy way, taking into consideration the unique class count of each dataset. Specifically, we observe that selecting 4-10 segments based on the class count in each dataset enhances performance. Consequently, we use an equal number of segments in each dataset for segment-based poolings (Appendix. Table \ref{tab: optimal parameter}). 

\textbf{Baselines}
We conduct two experiments to evaluate the performance of SoM-TP. First, we compare it with traditional temporal poolings, GTP, STP, and DTP, to demonstrate the effectiveness of selection-ensemble in temporal poolings \citep{lee2021learnable}. Second, we compare SoM-TP with other state-of-the-art models that utilize advanced methods, including scale-invariant methods (ROCKET \citep{dempster2020rocket}, InceptionTime \citep{ismail2020inceptiontime}, OS-CNN \citep{tang2020omni}, and DSN \citep{xiao2022dynamic}), sequential models (MLSTM-FCN \citep{karim2019multivariate}, and TCN \citep{bai2018empirical}), and Transformer-based models (Vanilla-Transformer \citep{vaswani2017attention}, TST \citep{zerveas2021transformer}, and ConvTran \cite{ConvTran2023}). Models leveraging temporal information use attention or RNNs to emphasize long-term dependencies. On the other hand, scale-invariant learning models employ a CNN-based architecture with various kernel sizes to find the optimum through global average pooling.

\subsection{Experimental Evaluation}
\label{sec4.2}

\subsubsection{Performance Analysis}
As shown in Table \ref{main:t2}, SoM-TP shows superior performance for overall TSC datasets when compared to conventional temporal poolings. We calculate the average performance of the entire repository. We consider not only accuracy but also the F1 macro score, ROC-AUC, and PR-AUC to consider the imbalanced class. Quantitatively, SoM-TP outperforms the existing temporal pooling methods both in univariate and multivariate time series datasets. Through these results, we can confirm that the dynamic selection ensemble of SoM-TP boosts the performance of the CNN model.

Additionally, Table \ref{tab:2} compares SoM-TP with other state-of-the-art TSC models from two different approaches. In Table \ref{tab:2}-1, ResNet SoM-TP MAX significantly outperforms other sequential models in terms of comparing models leveraging temporal information. As SoM-TP clearly outperforms all other models in accuracy metric, we demonstrate the robustness of performance by providing the number of datasets where SoM-TP achieves higher accuracy. Considering the lowest average rank of SoM-TP, we can conclude that dynamic pooling selection leverages the model to keep important temporal information in a more optimal way than other methods in the massive UCR/UEA repository.

Next, Table \ref{tab:2}-2 highlights SoM-TP’s comparable performance alongside scale-invariant methods, even with SoM-TP’s significant computational efficiency. Since SoM-TP and scale-invariant methods have different learning approaches, it is suitable to consider various metrics. Regarding the time complexity of models, scale-invariant methods consider various receptive fields of a CNN, which results in longer training times. In contrast, SoM-TP achieves comparable performance with only one-third of the time.

\begin{figure*}[h]
\centering
    \subfloat[acc: 0.9219]{\includegraphics[width=0.23\textwidth]{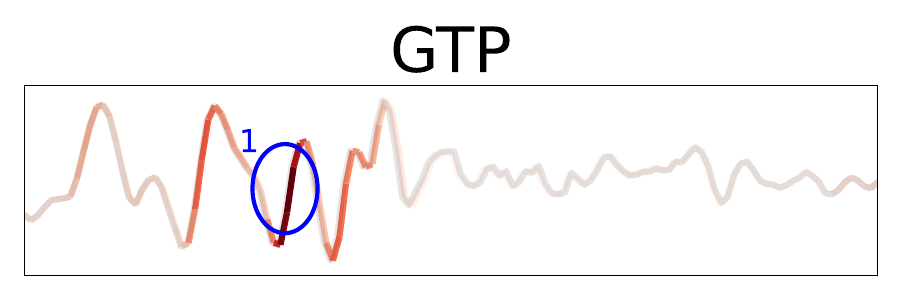}}
    \subfloat[acc: 0.7893]{\includegraphics[width=0.23\textwidth]{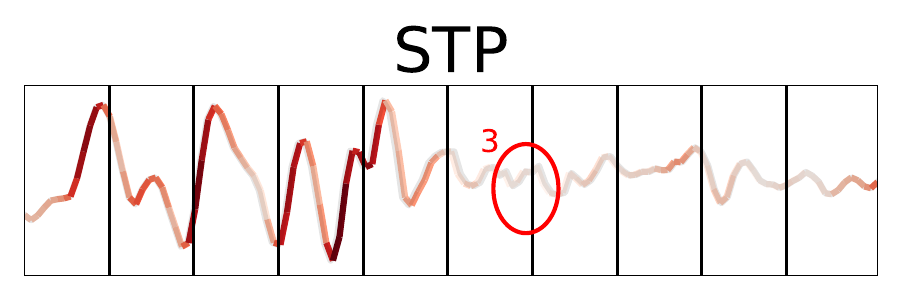}}
    \subfloat[acc: 0.8988]{\includegraphics[width=0.23\textwidth]{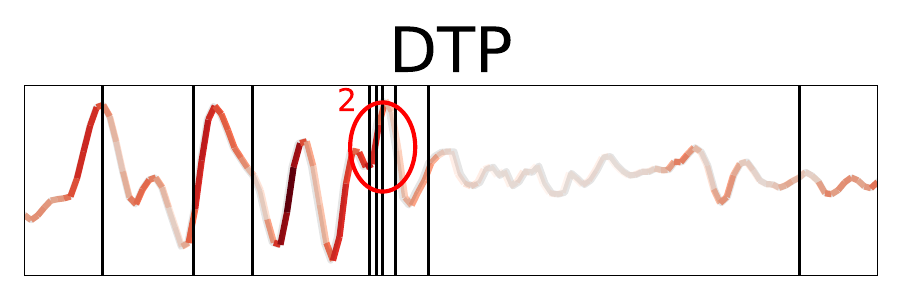}}
    \subfloat[acc: \bf 0.9396]{\includegraphics[width=0.23\textwidth]{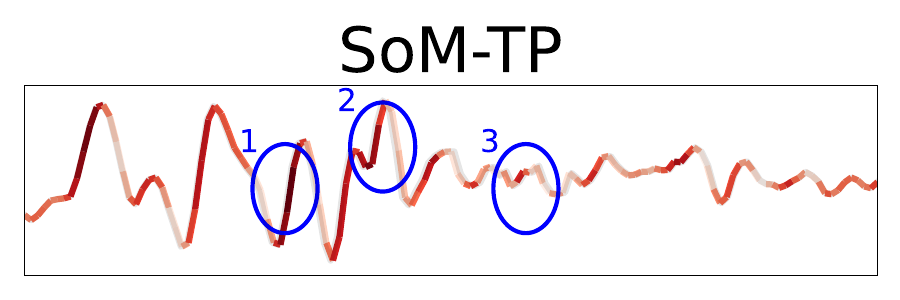}}\\

    \subfloat[acc: 0.3143]{\includegraphics[clip, trim=0cm 0cm 0cm 1.3cm, width=0.23\textwidth]{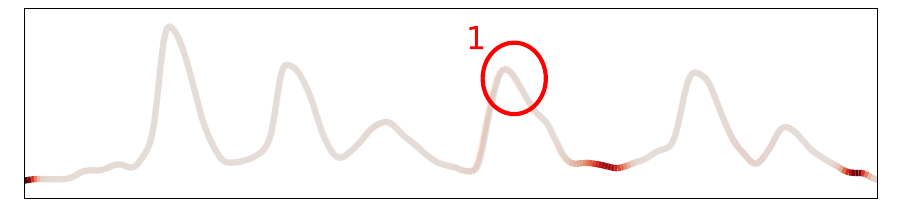}}
    \subfloat[acc: 0.6769]{\includegraphics[clip, trim=0cm 0cm 0cm 1.3cm, width=0.23\textwidth]{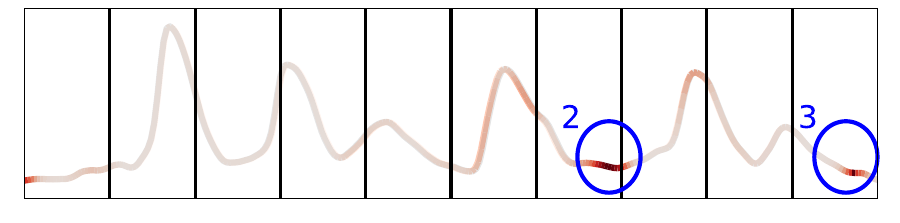}}
    \subfloat[acc: 0.5429]{\includegraphics[clip, trim=0cm 0cm 0cm 1.3cm, width=0.23\textwidth]{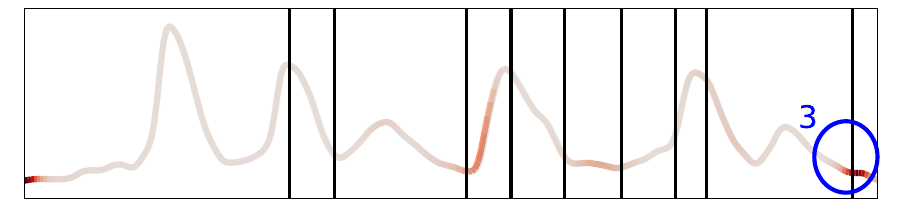}}
    \subfloat[acc: \bf 0.6901]{\includegraphics[clip, trim=0cm 0cm 0cm 1.3cm, width=0.23\textwidth]{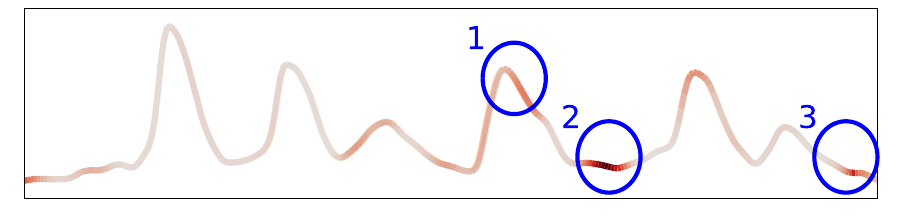}}\\

    \subfloat[acc: 0.3914]{\includegraphics[clip, trim=0cm 0cm 0cm 1.3cm, width=0.23\textwidth]{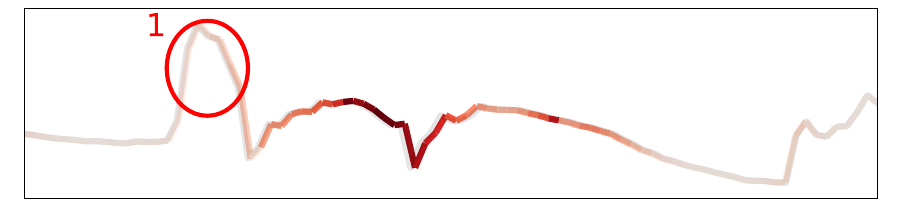}}   
    \subfloat[acc: 0.8139]{\includegraphics[clip, trim=0cm 0cm 0cm 1.3cm, width=0.23\textwidth]{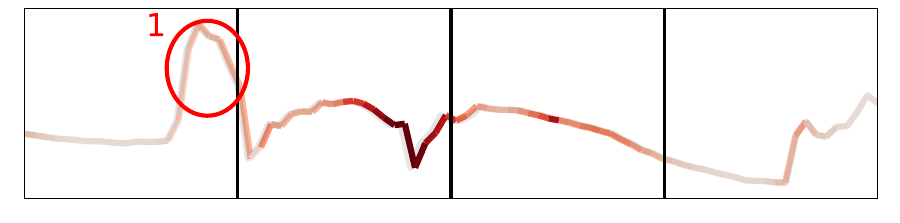}}
    \subfloat[acc: 0.7780]{\includegraphics[clip, trim=0cm 0cm 0cm 1.3cm, width=0.23\textwidth]{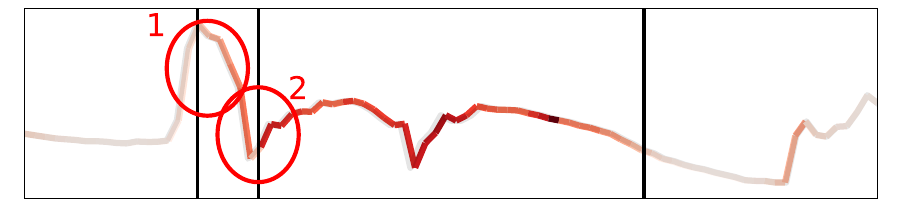}}
    \subfloat[acc: \bf 0.8810]{\includegraphics[clip, trim=0cm 0cm 0cm 1.3cm, width=0.23\textwidth]{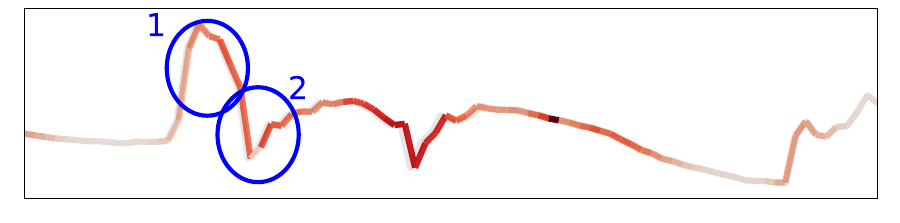}}
    
    \caption{Comparison of LRP Input Attribution on Single vs Diverse Perspective Learning. The figure shows LRP attribution results for FaceAll, FiftyWords, and MoteStrain datasets in the UCR repository, ordered in respective rows. Pooling choice significantly affects accuracy and attributions, reflecting different perspectives. Redder areas of time series indicate higher attribution, aligning with LRP's conservation rule of summing to 1. Blue circles denote well-captured regions, while red circles suggest dispersed focus or inadequate capture. Given the absence of a ground truth concept for input attributes in TSC, we infer these implicitly from the presented accuracy.} 
    \label{fig:mesh4}
\end{figure*}

Finally, in Table \ref{tab:small}, we present the results of an ablation study on the modules of SoM-TP discussed in Section 3. When each module, including $\mathbf{A}_0$ and $\phi_0$ constituting DPL Attention, DPLN, and $\mathcal{L}_{attn}$, is removed, it decreases SoM-TP's performance. In terms of dataset robustness, we can observe through the rank results that all modules contribute to promoting SoM-TP's Diverse Perspective Learning.

\subsubsection{Perspective Analysis with LRP}

In Figure \ref{fig:3}, we can observe that SoM-TP dynamically selects pooling during inference. ArrowHead, Chaintown, and ACSF1, in order, are datasets where GTP, STP, and DTP pooling selections are most frequently chosen. The DPL attention is trained to select the optimal pooling for each batch during the training process, and during inference, it continues to choose the most suitable pooling without DPLN (Appendix. Figure \ref{fig:extend3}).

For qualitative analysis, we employ Layer-wise Relevance Propagation (LRP) to understand how different temporal pooling perspectives capture time series patterns. LRP attributes relevance to input features, signifying their contribution to the output. Note that the conservation rule maintains relevance sum in backward propagation, ensuring that the sum of attribution is 1. We use the LRP $z^+$ rule for the convolutional stack $\Phi$, and the $\epsilon$ rule for the FC layers.

 \begin{table}[t]
     \centering
     \begin{adjustbox}{width=0.4\textwidth}
     \begin{tabular}{c|c|c}
     \hline
     \multirow{2}{*}{Model} & \multicolumn{2}{c}{Complexity} \\
     \cline{2-3}
      & Pooling & Optimization \\
     \hline
          GTP & $\mathcal{O}(1)$ & \multirow{3}{*}{$\mathcal{O}(N)$}\\
          STP/DTP & $\mathcal{O}(L)$ &  \\
          SoM-TP & $\mathcal{O}(L_\mathbf{P}) + \mathcal{O}(L_{mul}) = \mathcal{O}(L)$ & \\
         \hline
         MCL & - & $\mathcal{O}(N^2)$\\
     \hline
    \end{tabular}
     \end{adjustbox}
     \caption{Complexity Study.}
     \label{tab:complexity}
 \end{table}
 
In Figure \ref{fig:mesh4}, GTP focuses on globally crucial parts (a, e, i), while STP and DTP use local views within segmented time series for a more balanced representation (b, c, f, g, j, k). However, GTP's limitation lies in concentrating only on specific parts, neglecting other local aspects (e, i). Conversely, STP and DTP risk diluting the primary representation by reflecting all local segments (b, j) or cutting significant series due to forced segmentation (c, k). DTP often segments at change points, losing essential information (c, k).

SoM-TP addresses these issues by combining global and local views of each pooling method via diverse perspective learning. In Figure \ref{fig:mesh4}-(d), SoM-TP captures GTP's points (d-1) and enhances multiple representations by effectively capturing local patterns (d-2, d-3). In (h), SoM-TP identifies the common important points (h-2, h-3) and complements GTP's missed local points (h-1). Finally, in (l), SoM-TP captures GTP's missed local points (l-1) and fully utilizes important time series (l-2) cut by STP and DTP (j-1, k-1, k-2). 

\subsubsection{Complexity of SoM-TP} 
We compare the complexity of independent temporal poolings: pooling and optimization complexity. We exclude the maximum or average operation, which is common for all pooling complexity. 

As shown in Table \ref{tab:complexity}, for the pooling complexity, GTP has $\mathcal{O}(1)$ while STP and DTP have $\mathcal{O}(L)$ from segmenting. SoM-TP has increased complexity as $\mathcal{O}(L_\mathbf{P} + L_{mul}) = \mathcal{O}(L)$ for computation of the attention score, where $\mathcal{O}(L_\mathbf{P})$ is for a sum of the three temporal poolings’ complexity, making it $\mathcal{O}(L)$, and $\mathcal{O}(L_{mul})$ for the complexity of multiplication between $\Bar{\mathbf{P}}$ and $\mathbf{A_0}$, and between $\Bar{\mathbf{P}}$ and $\mathbf{A}$.
As for the optimization complexity, SoM-TP and other temporal poolings have all $\mathcal{O}(N)$, while MCL has $\mathcal{O}(N^2)$ to generate and compare multiple outputs. Therefore, compared with independent pooling, SoM-TP has little degradation of complexity, while optimization is effectively achieved even with an ensemble.

\section{Conclusion}
\label{sec5}
This paper proposes SoM-TP, a novel temporal pooling method employing a selection ensemble to address data dependency in temporal pooling by learning diverse perspectives. Utilizing a selection ensemble inspired by MCL, SoM-TP adapts to each data batch's characteristics. Optimal pooling selection with DPL attention achieves a comparison-free ensemble. We define DPLN and perspective loss for effective ensemble optimization. In quantitative evaluation, SoM-TP surpasses other pooling methods and state-of-the-art TSC models in UCR/UEA experiments. In qualitative analysis, LRP results highlight SoM-TP’s ability to complement existing temporal pooling limitations. We re-examine the conventional role of temporal poolings, identify their limitations, and propose an efficient data-driven temporal pooling ensemble as a first attempt.

\clearpage
\section{Acknowledgements}
This work was partly supported by Institute of Information \& Communications Technology Planning \& Evaluation (IITP) grant funded by the Korea government (MSIT) (No. 2022-0-00984, Development of Artificial Intelligence Technology for Personalized Plug-and-Play Explanation and Verification of Explanation; No. 2022-0-00184, Development and Study of AI Technologies to Inexpensively Conform to Evolving Policy on Ethics; No. 2021-0-02068, Artificial Intelligence Innovation Hub; No. 2019-0-00075, Artificial Intelligence Graduate School Program (KAIST)).

\bibliography{aaai24_2}

\clearpage
\appendix
\section{Related Work}
\subsection{Time Series Classification Methods}
In recent Time Series Classification (TSC) tasks, deep learning models have shown significant performance improvement \cite{lee2021learnable,wang2017time,ismail2019deep}, surpassing that of traditional machine learning models such as DTW \cite{rakthanmanon2012searching}, BOSS \cite{schafer2015boss}, COTE \cite{7069254}, and HIVE-COTE \cite{7837946}. Various deep learning methodologies have been proposed to address the characteristics of time series, broadly classified into two categories: 1) learning diverse receptive fields using various convolution kernels to handle variant scales of time series, and 2) preserving temporal information to reduce the loss of unique information in time series.

\subsubsection{Scale-Invariant Learning Methods}
Models belonging to scale-invariant methods, such as MCNN \cite{cui2016multi}, InceptionTime \cite{ismail2020inceptiontime}, Tapnet \cite{zhang2020tapnet}, ROCKET \cite{dempster2020rocket,multirocket}, OS-CNN \cite{tang2020omni}, and DSN \cite{xiao2022dynamic}, utilize various convolutional kernels to cover multiple receptive fields and extract relevant features across different time series scales. The incorporation of scale-invariant properties in these models has significantly contributed to the improvement of TSC performance. However, these models are limited to convolutional design, have high computational costs, and can be easily overfitted, particularly when trained on small datasets. In general, finding an adaptive receptive field largely depends on optimal kernel size or masks during training, a process that is sparse and imposes substantial computational costs \cite{xiao2022dynamic}.

\subsubsection{Temporal Information Leveraging  Methods}
Recent research has focused on scale-invariant learning through the application of diverse convolution kernels, achieving remarkable results. Despite this progress, we still emphasize the importance of utilizing another crucial feature of time series data: temporal information \cite{lee2021learnable,zhao2022t}. One approach to reflect this is to use the CNN structure for TSC, as proposed by \citet{wang2017time}, along with FCN and ResNet. These architectures stack CNN layers without pooling to preserve temporal information. However, this information is ultimately lost when global pooling is used in the final layer. Alternatively, some models directly utilize RNN layers, such as MLSTM-FCN \cite{karim2019multivariate}, use a Transformer-based approach like TST \cite{zerveas2021transformer}, or redesign convolution for sequential modeling, such as TCN \cite{bai2018empirical}. However, sequential models have higher complexity than CNN-based models and are limited to performing well only on time series data with long-term dependencies.

\subsection{Advanced Temporal Pooling Methods}
In CNNs, global pooling is commonly used to reduce dimensionality and computation costs. One of the most frequently used methods is MAX and AVG global pooling, which has demonstrated great success in classification tasks. However, it has been criticized for its suboptimal information preservation \cite{sabour2017dynamic, hinton2018matrix}. This drawback can be illustrated in two ways. First, global pooling causes a significant decrease in dimensionality, leading to a loss of information in the data. Second, when the MAX operation is used, global pooling only reflects local information and constrains the ability to effectively minimize approximation loss \cite{rippel2015spectral}.

To address the issue in time series data, \citet{lee2021learnable} proposed Dynamic Temporal Pooling (DTP) to mitigate the temporal information loss caused by global pooling in the CNN architecture designed by \citet{wang2017time}. DTP employs a soft-DTW \cite{cuturi2017soft} layer to cluster similar time segments before pooling. DTP relies on aligning hidden features temporally but imposes a constraint against aligning a single time point with multiple consecutive segments. This constraint in segmentation may lead to the inadvertent separation of crucial change points that are vital for preserving time series patterns. Additionally, another advanced temporal pooling method, MultiRocket \cite{multirocket}, is tailored specifically to be efficient and scalable within the ROCKET architecture \cite{dempster2020rocket}. Therefore, it poses challenges when attempting to apply it to the pooling layer within a general CNN model architecture.

The pursuit of advanced pooling techniques extends beyond Time Series Classification (TSC) and encompasses diverse domains. For example, researchers are actively exploring ways to preserve image sequence information in pooling for video data \cite{gao2021temporal,girdhar2017attentional,pigou2018beyond,wang2021spatial}. Similarly, in the image domain, ongoing research aims to maintain spatial information during pooling \cite{alsallakhmind,hou2020strip}. It is noteworthy that minimizing information loss during the pooling process remains a significant challenge across these diverse domains as well.

\subsection{Difference between SoM-TP and Neural Architecture Search}
Neural Architecture Search (NAS) \cite{zoph2017neural} is a methodology designed to automatically discover optimal or highly effective neural network architectures tailored for specific tasks. In contrast to NAS, which aims to find the most efficient configuration for all modules in the entire model architecture, SoM-TP has a distinct objective: the selection of optimal pooling methods. Even when applying NAS only to the pooling layer, SoM-TP possesses two notable differences from NAS \cite{liu2019darts}. First, SoM-TP dynamically selects the optimal pooling path for each data batch, while NAS converges towards a fixed path based solely on overall dataset performance. Second, in terms of optimization, NAS faces a bilevel optimization problem to identify the path for the entire network, while SoM-TP optimizes a single classifier by utilizing the sub-network (DPLN) loss as a regularizer. In summary, SoM-TP has clear differences from NAS in terms of convergence and optimization.

\onecolumn
\section{Extended Analysis on SoM-TP}
\label{appendix1}

\begin{figure*}[h]
    \centering
    \includegraphics[width=0.9\textwidth]{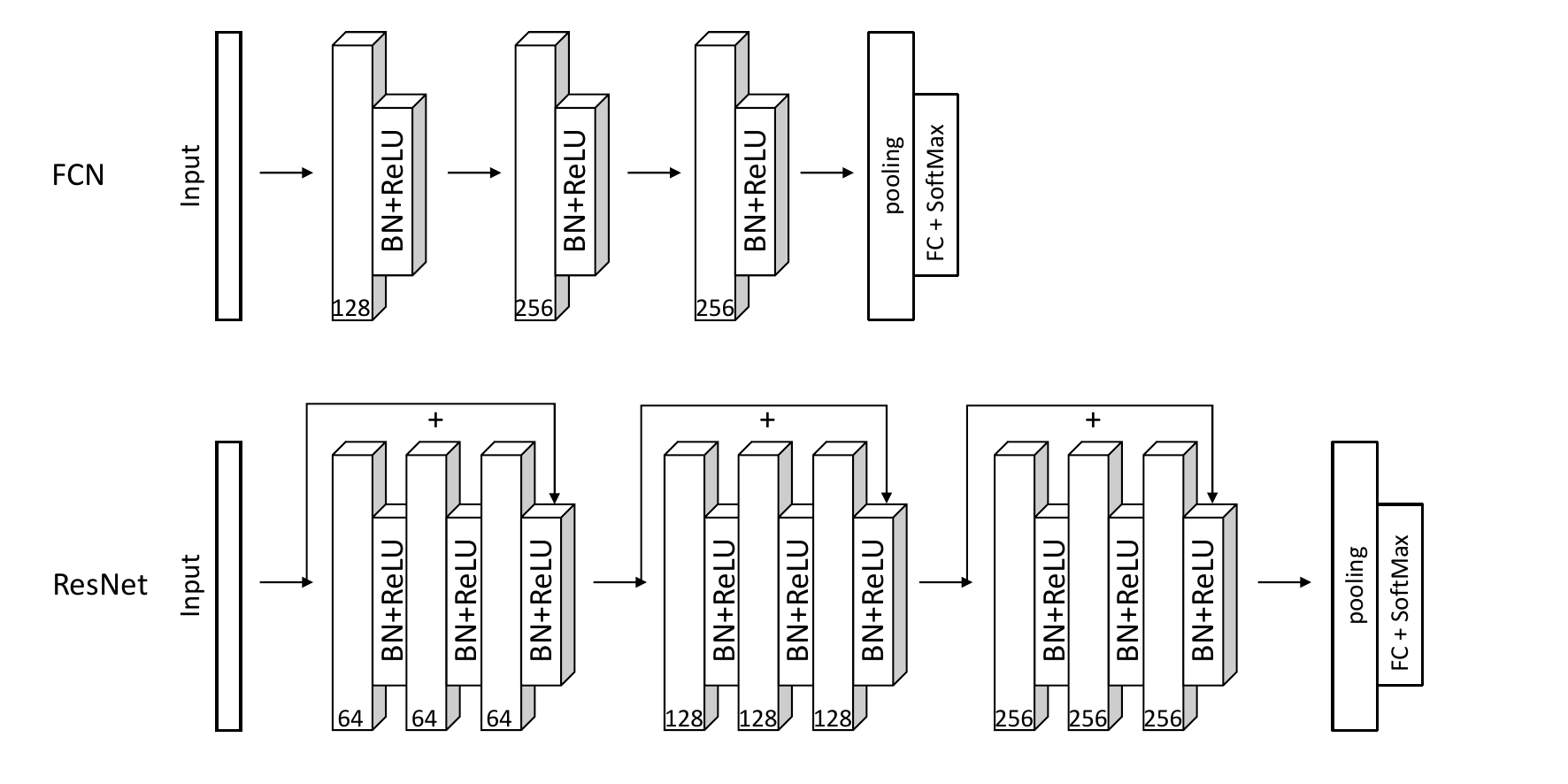}
    \caption{\textbf{Convolutional Stack Architectures.} The overall architectures of FCN and ResNet used in this paper follow the baseline models presented by \citet{wang2017time}, specifically designed for Time Series Classification (TSC). In both architectures, a convolutional layer preceding pooling and FC layers for classification decisions are implemented. FCN consists of three convolutional layers, with hidden dimension sizes sequentially set at 128, 256, and 256, and kernel sizes of 9, 5, and 3. ResNet comprises nine convolutional layers, with hidden dimension sizes sequentially set as 64 $\times$ 3, 128 $\times$ 3, 256 $\times$ 3, and kernel sizes of (9, 5, 3) $\times$ 3. SoM-TP is applied to the single pooling layer just before the FC layer in both model architectures. The FC layers of the CLS network and DPLN consist of three linear layers each. The FC layers of the CLS network are structured as follows: 256 * number of segments $\times$ 512, 512 $\times$ 1024, 1024 $\times$ number of classes. On the other hand, the FC layers of the DPLN are structured as: 256 * number of segments * 3 $\times$ 512, 512 $\times$ 1024, 1024 $\times$ number of classes.}
    \label{fig:app1.1}
\end{figure*}
\subsubsection{Model Architecture Choice}
SoM-TP is a pooling method that fully leverages temporal information from the convolutional encoder. This implies that SoM-TP requires the utilization of the preserved information from the convolutional encoder. Consequently, an architecture capable of accumulating detailed information in deeper layers, such as ResNet, is suitable for SoM-TP. When SoM-TP replaces the final global average pooling layer in ResNet, it surpasses the performance of the vanilla ResNet.

Technically, SoM-TP can be implemented in intermediate pooling layers of models with multiple pooling stages. However, considering SoM-TP's objective of selecting the optimal temporal pooling that fully reflects the formed representations, applying it to the accumulated information across all layers in the last layer is the most effective approach.

\subsubsection{Detailed Training and Evaluation Procedures of SoM-TP}
The main distinction between training and evaluation is that the DPLN subnetwork is not used during evaluation. In the evaluation process, only the trained $\mathbf{A}_0$ and $\phi_0$ are used to select the optimal pooling for each batch, leading to the generation of the output $\mathbf{A}$. Subsequently, the pooling output is selected from $\mathbf{A}$.

\clearpage
\subsubsection{Optimal Hyper-parameters}
\begin{table}[h]
    \centering
    \begin{adjustbox}{width=0.6\textwidth}
    \begin{tabular}{cccccccccc}
        \noalign{\smallskip}\noalign{\smallskip}\hline
        \bf Network & \bf \#Conv. & \multicolumn{2}{c}{\bf \# Segments} & \bf \#FC. & \bf Optim. & \bf lr & \bf batch & \bf window & \bf epoch\\

            & & GTP  & STP\&DTP &\\
        \hline
        FCN & 3 & \multirow{2}{*}{1} & \multirow{2}{*}{$\mathbf{n}$} & \multirow{2}{*}{3} & \multirow{2}{*}{Adam} & \multirow{2}{*}{$1e-4$} & \multirow{2}{*}{[$1, 8$]} & \multirow{2}{*}{$1$} & \multirow{2}{*}{$300$} \\ 
        
        ResNet & 9 & & & & \\ 
        \hline
    \end{tabular}
    \end{adjustbox}
    
    \begin{adjustbox}{width=0.6\textwidth}
    \begin{tabular}{c|c|ccc|c}
    \hline
         Data & Type & GTP & STP & DTP & $\lambda$ decay \\
         \hline
         \multirow{4}{*}{UCR} & FCN SoM-TP MAX & \multirow{2}{*}{avg} & \multirow{2}{*}{avg} & \multirow{2}{*}{max} & \multirow{2}{*}{$1e-1$} \\
         & ResNet SoM-TP MAX & & & & \\
        \cline{2-6}
        &FCN SoM-TP AVG & avg & avg & max & \multirow{2}{*}{$1.0$}\\
        &ResNet SoM-TP AVG & max & max & max &\\
        \hline 
        \multirow{4}{*}{UEA}& FCN SoM-TP MAX & max & max & max & $1e-1$\\
        & ResNet SoM-TP MAX & max & max & avg & $1e-2$\\
        \cline{2-6}
        &FCN SoM-TP AVG & max & max & max & $1e-1$\\
        &ResNet SoM-TP AVG & avg & max & max & $5e-3$ \\
    \hline
    \end{tabular}
    \end{adjustbox}
    \caption{Optimal Hyper-parameters.}
    \label{tab: optimal parameter}
\end{table}

\begin{table*}[!ht]
\centering
\begin{adjustbox}{width=0.95\textwidth}
        \begin{tabular}{ll|ll|ll|ll|ll|ll|ll}
             \multicolumn{2}{c|}{GTP} & \multicolumn{2}{c|}{STP} & \multicolumn{2}{c|}{DTP} & \multicolumn{2}{c|}{FCN-UCR} & \multicolumn{2}{c|}{FCN-UEA} & \multicolumn{2}{c|}{Res-UCR} & \multicolumn{2}{c}{Res-UEA}\\
             MAX & AVG & MAX & AVG & MAX & AVG & ACC & F1 & ACC & F1 & ACC & F1 & ACC & F1\\
             \hline
             \multirow{4}{*}{\checkmark} & & \multirow{2}{*}{\checkmark} & & \checkmark & & 0.7502  & 0.7195  & \bf 0.6920 & \bf 0.6621 & 0.7600 & 0.7319 & 0.6453 & 0.6187 \\
             \cline{5-14}
             & &  & & & \checkmark & 0.7351 & 0.7062 & 0.6849 & 0.6554 & 0.7533 & 0.7214 & \bf 0.6769 & \bf 0.6387\\
             \cline{3-14}
             & & & \multirow{2}{*}{\checkmark} & \checkmark & & 0.7508 & 0.7223 & 0.6611 & 0.6439 & 0.7652 & 0.7382 & 0.6610 & 0.6300\\
             \cline{5-14}
             & & & & & \checkmark & 0.7425 & 0.7125 & 0.6867 & 0.6590 & 0.7607 & 0.7318 & 0.6579 & 0.6325 \\
             \cline{1-14}
             & \multirow{4}{*}{\checkmark} & \multirow{2}{*}{\checkmark} & & \checkmark & & 0.7515 & 0.7183 & 0.6644 & 0.6406 & 0.7628 & 0.7323 & 0.6598 & 0.6217 \\
             \cline{5-14}
             & &  & & & \checkmark & 0.7370 & 0.6989 & 0.6707 & 0.6430 & 0.7713 & 0.7432 & 0.6516 & 0.6242 \\
             \cline{3-14}
             & & & \multirow{2}{*}{\checkmark} & \checkmark & & \bf 0.7556 & \bf 0.7241 & 0.6848 & 0.6633 & \bf 0.7773 & \bf 0.7489 & 0.6734 & 0.6399\\
             \cline{5-14}
             & & & & & \checkmark & 0.7398 & 0.7055 & 0.6539 & 0.6383 & 0.7678 & 0.7396 & 0.6573 & 0.6299 \\
             \hline
             \multicolumn{8}{l}{\small *This table is for SoM-TP pooling selection operation type MAX.} \\
             \\
        \end{tabular}
    \end{adjustbox}
    \bigskip
    \begin{adjustbox}{width=0.95\textwidth}
        \begin{tabular}{ll|ll|ll|ll|ll|ll|ll}
             \multicolumn{2}{c|}{GTP} & \multicolumn{2}{c|}{STP} & \multicolumn{2}{c|}{DTP} & \multicolumn{2}{c|}{FCN-UCR} & \multicolumn{2}{c|}{FCN-UEA} & \multicolumn{2}{c|}{Res-UCR} & \multicolumn{2}{c}{Res-UEA}\\
             MAX & AVG & MAX & AVG & MAX & AVG & ACC & F1 & ACC & F1 & ACC & F1 & ACC & F1\\
             \hline
             \multirow{4}{*}{\checkmark} & & \multirow{2}{*}{\checkmark} & & \checkmark & & 0.7407  & 0.7147  & \bf 0.6909 & \bf 0.6753 & \bf 0.7659 & \bf 0.7389 & 0.6715 & 0.6523 \\
             \cline{5-14}
             & &  & & & \checkmark & 0.7278 & 0.6984 & 0.6445 & 0.6169 & 0.7526 & 0.7239 & 0.6750 & 0.6270\\
             \cline{3-14}
             & & & \multirow{2}{*}{\checkmark} & \checkmark & & 0.7347 & 0.7084 & 0.6794 & 0.6412 & 0.7508 & 0.7200 & 0.6674 & 0.6249\\
             \cline{5-14}
             & & & & & \checkmark & 0.7193 & 0.6894 & 0.6295 & 0.5982 & 0.7284 & 0.7012 & 0.6294 & 0.6050 \\
             \cline{1-14}
             & \multirow{4}{*}{\checkmark} & \multirow{2}{*}{\checkmark} & & \checkmark & & 0.7469 & 0.7171 & 0.6295 & 0.6804 & 0.7476 & 0.7232 & \bf 0.6764 & \bf 0.6505 \\
             \cline{5-14}
             & &  & & & \checkmark & 0.7136 & 0.6825 & 0.6563 & 0.6291 & 0.7303 & 0.7032 & 0.6467 & 0.6081 \\
             \cline{3-14}
             & & & \multirow{2}{*}{\checkmark} & \checkmark & & \bf 0.7562 & \bf 0.7315 & 0.6800 & 0.6510 & 0.7437 & 0.7315 & 0.6722 & 0.6486  \\
             \cline{5-14}
             & & & & & \checkmark & 0.7198 & 0.6870 & 0.6531 & 0.6274 & 0.7373 & 0.7103 & 0.6602 & 0.6278 \\
             \hline
             \multicolumn{8}{l}{\small *This table is for SoM-TP pooling selection operation type AVG.} \\
        \end{tabular}
    \end{adjustbox}
    \label{t2}
    \caption{Ablation Study on Pooling Operation in the Pooling Block.SoM-TP selects pooling by the MAX/AVG index of attention $\mathbf{A}$, and this is referred to as SoM-TP's operation. However, operations for each pooling within the pooling block can also vary. In other words, the 'max' and 'avg' operations of GTP, STP, and DTP can be hyperparameters in SoM-TP. We investigate the performance through ablation studies, and the best performance is indicated in bold.}
\end{table*}
The optimal hyperparameters for the results in Tables \ref{main:t2} and \ref{tab:2} can be found in Table \ref{tab: optimal parameter}. The batch size ranges from 1 to 8, with a default of 8 or one-tenth of the dataset size if the dataset size is under 80. The pooling operation occurs once in both the pooling block and DPL Attention $\mathbf{A}$. In Table \ref{tab: optimal parameter}, 'Type' refers to the operation type (AVG, MAX) used when selecting pooling from DPL attention $\mathbf{A}$. The operation (avg, max) for GTP, STP, and DTP denotes the pooling operation employed within the pooling block. The $\lambda$ decay parameter represents the degree of influence of the perspective loss and is optimized within the range of [1e-5, 1.0]. The seed number is arbitrarily set to 10.

\clearpage
\begin{figure*}[h]
    \centering
    \subfloat[FCN SoM-TP in UCR]{\includegraphics[width=0.23\textwidth]{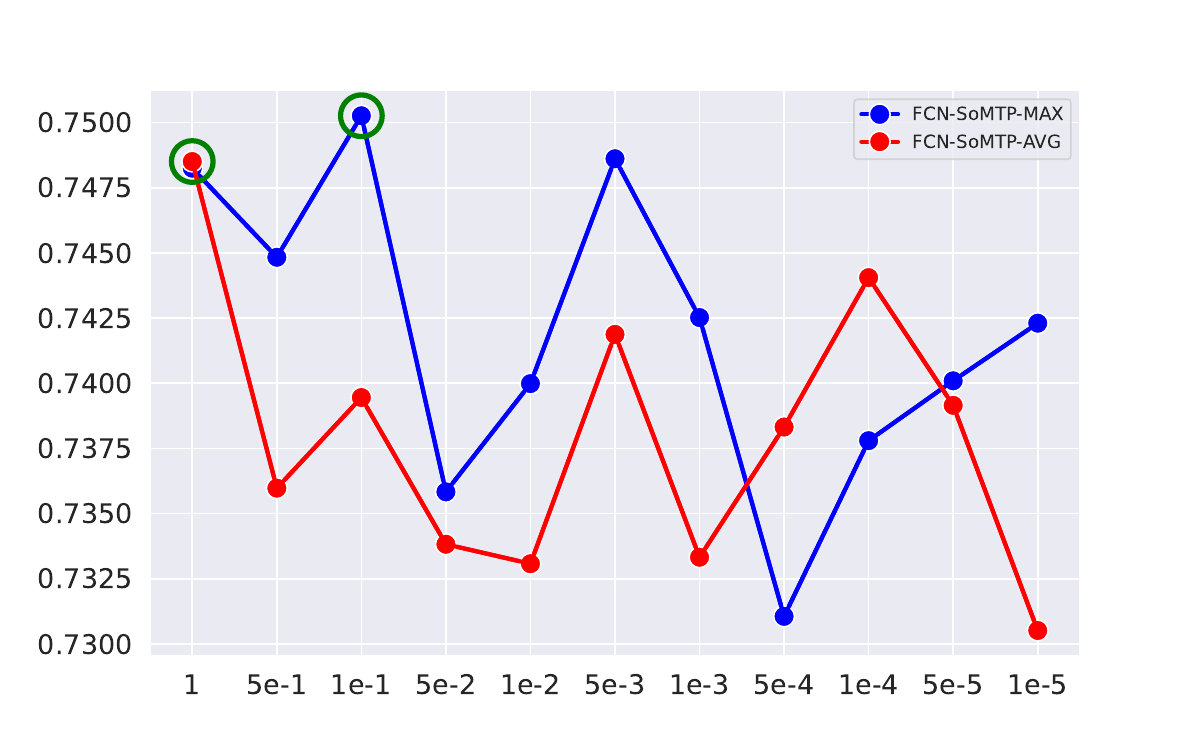}}
    \subfloat[FCN SoM-TP in UEA]{\includegraphics[width=0.23\textwidth]{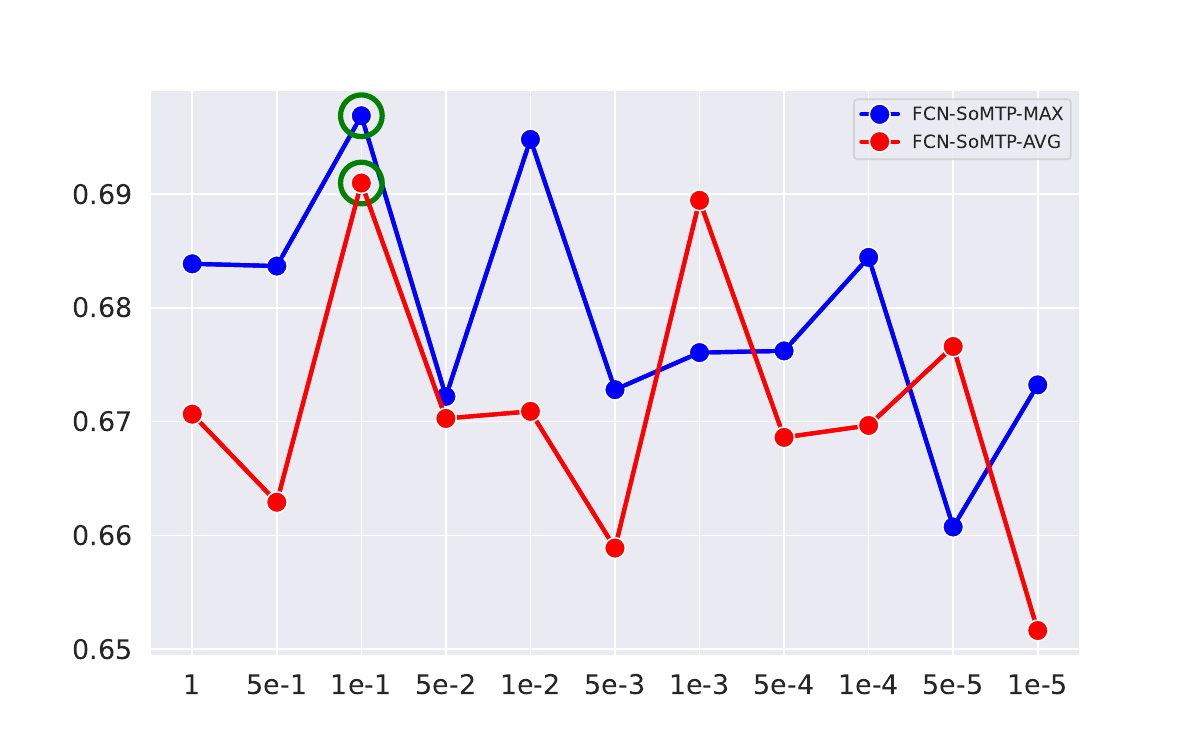}}
    \subfloat[ResNet SoM-TP in UCR]{\includegraphics[width=0.23\textwidth]{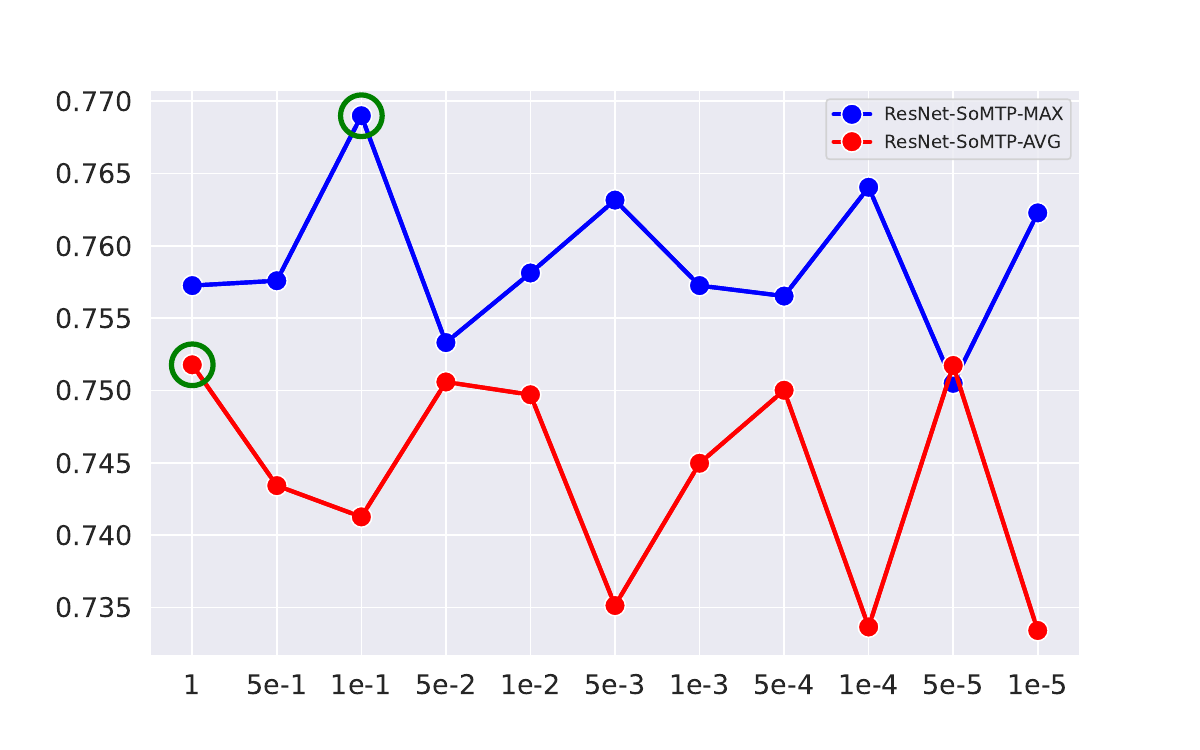}}
    \subfloat[ResNet SoM-TP in UEA]{\includegraphics[width=0.23\textwidth]{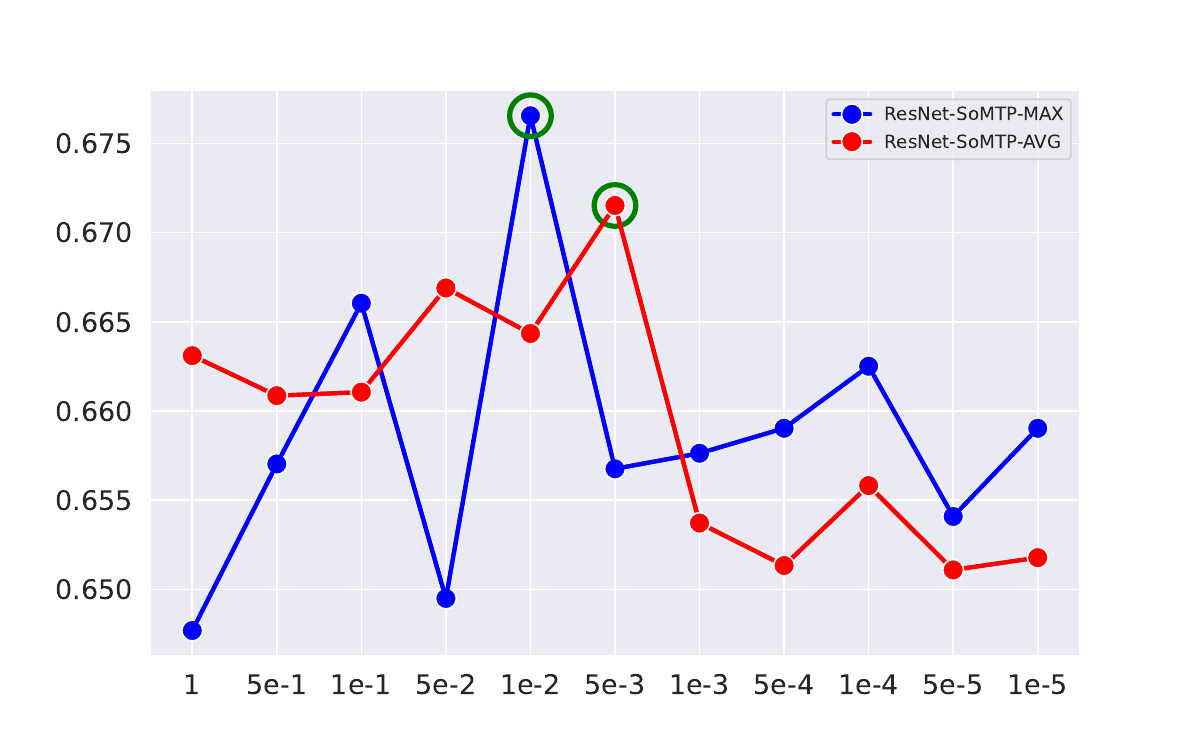}}\\
    \caption{$\lambda$ ablation study of SoM-TP. $\lambda$ is the decay value of the perspective loss, a crucial hyperparameter in SoM-TP. Note that as $\lambda$ increases, the perspective loss reflects all pooling perspectives more strongly. The ablation study covers $\lambda$ values across the range of [1, 1e-5] with 11 intervals. The blue and red lines correspond to the utilized pooling operation types, MAX and AVG, respectively. The optimal $\lambda$ resulting in the highest performance is highlighted with a green circle.}
    \label{fig:mesh6}
\end{figure*}
\subsubsection{How Does Perspective Loss Affect Pooling Ensemble?}
The degree of balanced pooling selection in SoM-TP is tunable by adjusting the similarity between the output distributions of $f_{DPLN}$ and $f_{CLS}$ using the perspective loss. The relationship between performance and the value of $\lambda$ is depicted in Figure \ref{fig:mesh6}. Increasing $\lambda$ facilitates the integration of multiple perspectives from diverse poolings in the ensemble via the perspective loss. The optimal balance depends on the dataset and model characteristics. Nevertheless, when assessed across various UCR/UEA datasets, Table \ref{tab:small} indicates that the absence of perspective loss (without DPLN) considerably diminishes the effectiveness of the selection ensemble.

\subsubsection{A Detailed Experiment on Batch}
SoM-TP selects the most proper pooling by minimizing the sum of losses for all samples within a batch. Therefore, we acknowledge that individual batches can have varying pooling selections based on their composition. However, since Attention $\mathbf{A}_0$ accumulates updates for each batch, it ultimately considers the entire dataset. Additionally, DPLN, through perspective loss, incorporates the results of other poolings not selected in the CLS network, making it robust to batch shuffling.

To confirm the robustness of SoM-TP to batch shuffling, we conduct experiments on 113 UCR datasets with FCN, using batch shuffling with 5 different seeds. The standard deviation of average performance for all datasets is $0.0046$, affirming its robustness to batch variations.

\subsection{Detailed Performance Analysis on SoM-TP}
\begin{table*}[!ht]
\centering
    \begin{adjustbox}{width=0.95\textwidth}
    \begin{tabular}{lll|lllll|lllll}
        \noalign{\smallskip}\noalign{\smallskip}
        \bf CNN & \multicolumn{2}{l}{\bf POOL (type)} & \multicolumn{5}{l}{\bf UCR (uni-variate)} & \multicolumn{5}{l}{\bf UEA (multi-variate)} \\
        & \multicolumn{2}{l|}{\bf AVG} & ACC & F1 macro & ROC AUC & PR AUC & Rank & ACC & F1 macro & ROC AUC & PR AUC & Rank \\
        \hline
        \multirow{5}{*}{FCN} & \multicolumn{2}{l|}{GTP} & 0.7227 & 0.6902 & \underline{0.8838} & 0.7641  & 2.8 & 0.6642 & 0.6388 
 & 0.8067 & 0.6974  & 3.0\\
                 & \multicolumn{2}{l|}{STP} & \underline{0.7302} & \underline{0.6967} & 0.8770 & \underline{0.7697} & \underline{2.9} &  \underline{0.6816} & \underline{0.6515} & \textbf{0.8171} & \underline{0.7119}  & \bf 2.6\\
                 & \multirow{2}{*}{DTP} & euc & 0.7137 & 0.6815 & 0.8735 & 0.7489  & 3.5 & 0.6449 & 0.6157 & 0.7722 & 0.6656  & 3.8 \\
                 & & cos & 0.7106 & 0.6774 & 0.8701 & 0.7470  & 3.5 & 0.6372 & 0.6088 & 0.7780 & 0.6612  & 3.9\\
                \cline{2-13}
                 & \multicolumn{2}{l|}{SoM-TP} & \bf 0.7562 & \bf 0.7315 & \bf 0.9016 & \bf 0.7957 & \bf 2.5 & \bf 0.6909 & \bf 0.6753 & \underline{0.8137} & \bf 0.7138  & \underline{2.7}\\
        \hline
        \multirow{5}{*}{ResNet} & \multicolumn{2}{l|}{GTP} & 0.7544 & 0.7244 & \textbf{0.9137} & \textbf{0.8070}  & 2.8 & 0.6419 & 0.6149 & 0.7845 & 0.6825  & 2.9\\
                 & \multicolumn{2}{l|}{STP} & \underline{0.7583} & \underline{0.7323} & 0.8983 & 0.7996  & \bf 2.6 & \underline{0.6612} & \underline{0.6321} & \underline{0.8022} & \underline{0.6950}  & \underline{2.9}\\
                 & \multirow{2}{*}{DTP} & euc & 0.7258 & 0.6967 & 0.8797 & 0.7649 & 3.4 & 0.6475 & 0.6227 & 0.7886 & 0.6859  & 3.3\\
                 & & cos & 0.7256 & 0.6963 & 0.8735 & 0.7563 & 3.4 & 0.6299 & 0.6044 & 0.7672 & 0.6540 & 3.9\\
                 \cline{2-13}
                 & \multicolumn{2}{l|}{SoM-TP} & \bf 0.7659 & \bf 0.7389 & \underline{0.9038} & \underline{0.8026}  & \underline{2.7} & \bf 0.6764 & \bf 0.6505 & \textbf{0.8081} & \bf 0.7105 & \bf 2.6 \\
        \hline
        \multicolumn{8}{l}{\small *This table is for SoM-TP pooling selection operation type AVG.} \\
    \end{tabular}
    \end{adjustbox}
    \label{t2}
    \caption{SoM-TP AVG Performance. Note that SoM-TP selects pooling by the MAX/AVG index of attention $\mathbf{A}$, referred to as SoM-TP's operation. This table represents SoM-TP selection operation AVG performance compared to existing temporal poolings. The best performances are bolded, and the second-best performances are underlined.}
\end{table*}

\clearpage
\subsection{Dynamic Selection of Temporal Poolings in SoM-TP}
\begin{figure*}[!ht]
\captionsetup[subfigure]{oneside,margin={0cm,3cm}}
\captionsetup[subfigure]{labelformat=empty}
    \subfloat[*GTP most selected]{\includegraphics[width=0.33\textwidth, clip, trim=0 1cm 0 0.3cm]{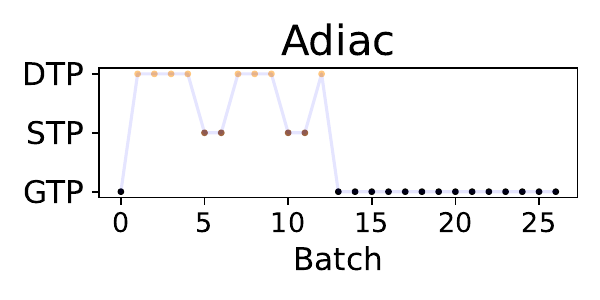}}
    \subfloat{\includegraphics[width=0.33\textwidth, clip, trim=0 1cm 0 0.3cm]{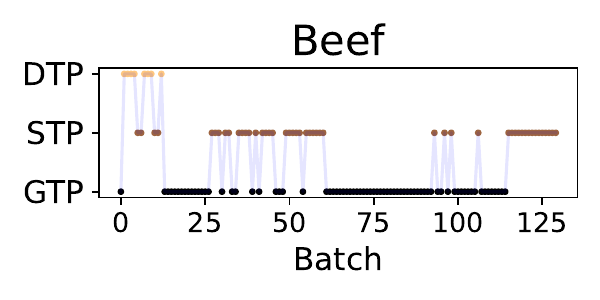}}
    \subfloat{\includegraphics[width=0.33\textwidth, clip, trim=0 1cm 0 0.3cm]{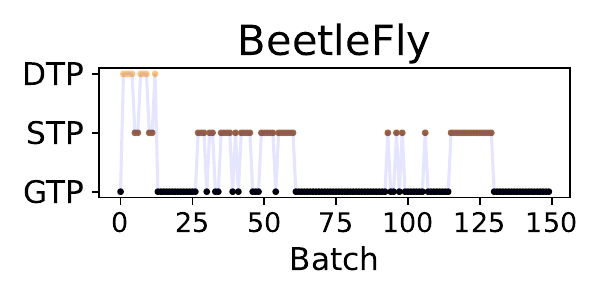}}\\

    \subfloat[*STP most selected]{\includegraphics[width=0.33\textwidth, clip, trim=0 1cm 0 0.3cm]{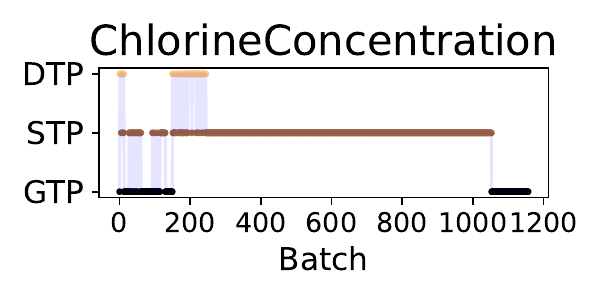}}
    \subfloat{\includegraphics[width=0.33\textwidth, clip, trim=0 1cm 0 0.3cm]{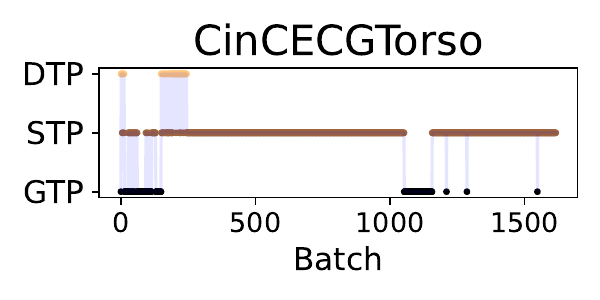}}
    \subfloat{\includegraphics[width=0.33\textwidth, clip, trim=0 1cm 0 0.3cm]{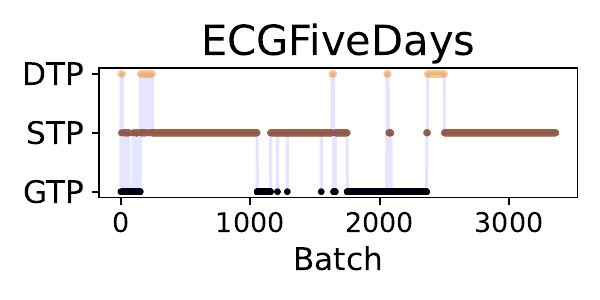}}\\

    \subfloat[*DTP most selected]{\includegraphics[width=0.33\textwidth, clip, trim=0 0.5cm 0 0.3cm]{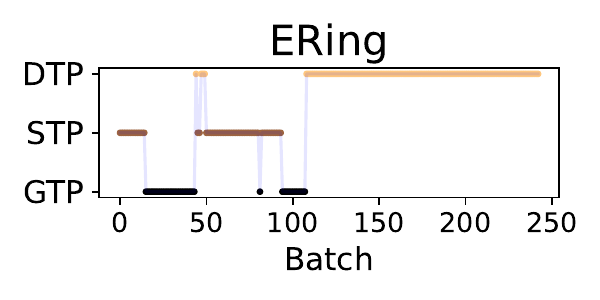}}
    \subfloat{\includegraphics[width=0.33\textwidth, clip, trim=0 0.5cm 0 0.3cm]{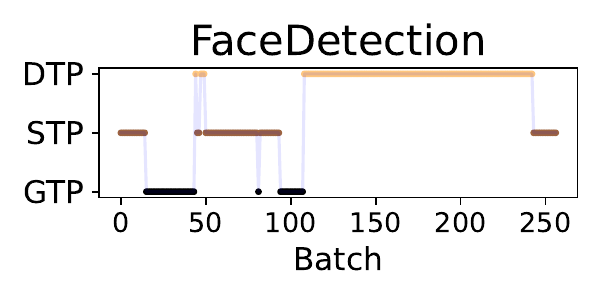}}
    \subfloat{\includegraphics[width=0.33\textwidth, clip, trim=0 0.5cm 0 0.3cm]{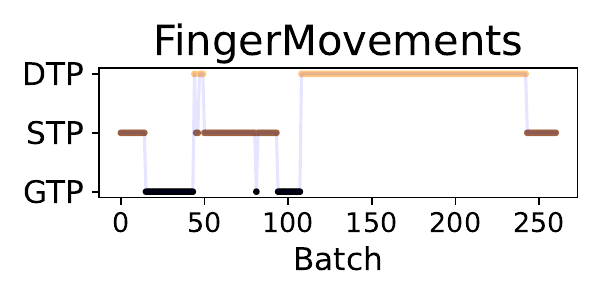}}

    \caption{Dynamic Pooling Selection in SoM-TP. Expanding on Figure \ref{fig:3}, this figure provides additional examples encompassing various time series data. Within this figure, we can examine the dynamic selection graph for FCN SoM-TP MAX on the UCR/UEA repository. We demonstrate that SoM-TP's dynamic selection mechanism functions robustly across multiple datasets with diverse data sizes and time lengths during the inference process. Notably, samples labeled as `DTP most selected' are chosen for datasets from the UEA repository, which typically present higher complexity compared to the UCR repository.}
    \label{fig:extend3}
\end{figure*}

\clearpage
\subsection{LRP Ablation Study on Different Temporal Poolings}
\begin{figure*}[h]
\centering
    \subfloat[acc: 0.7126]{\includegraphics[width=0.23\textwidth]{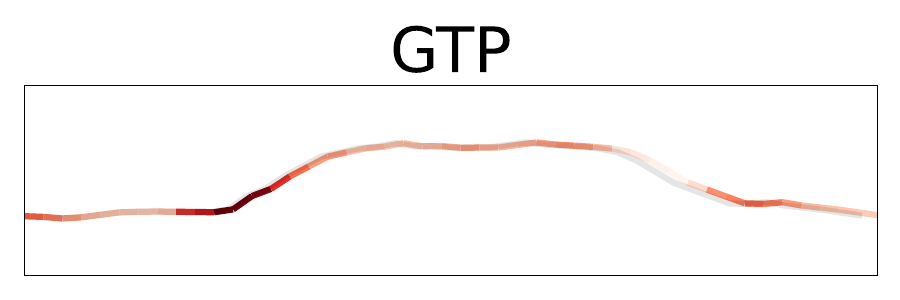}}
    \subfloat[acc: 0.7361]{\includegraphics[width=0.23\textwidth]{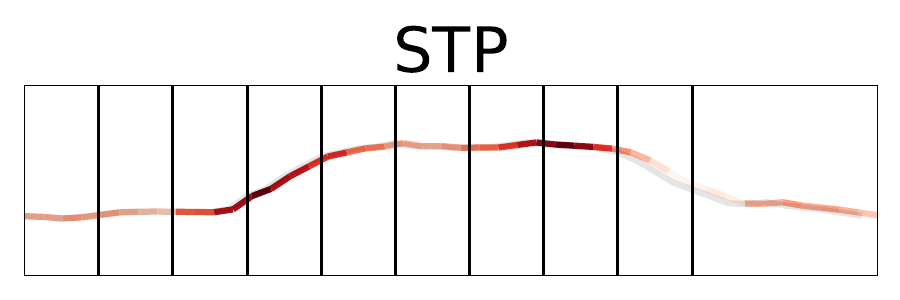}}
    \subfloat[acc: 0.7271]{\includegraphics[width=0.23\textwidth]{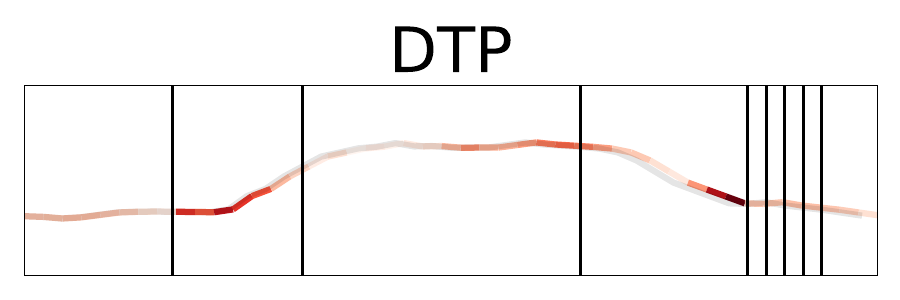}}
    \subfloat[acc: \bf 0.7549]{\includegraphics[width=0.23\textwidth]{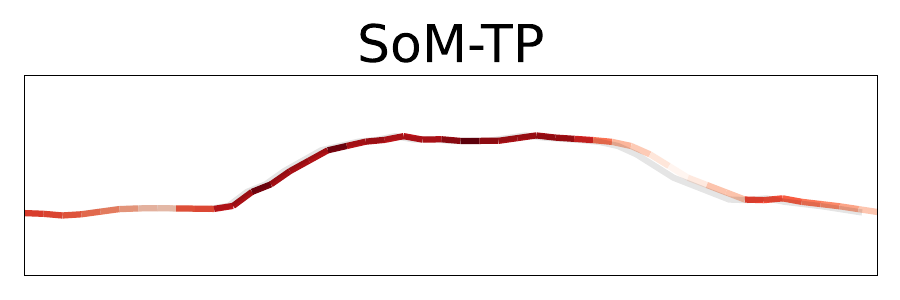}}\\

    \subfloat[acc: 0.9514]{\includegraphics[clip, trim=0cm 0cm 0cm 1.3cm, width=0.23\textwidth]{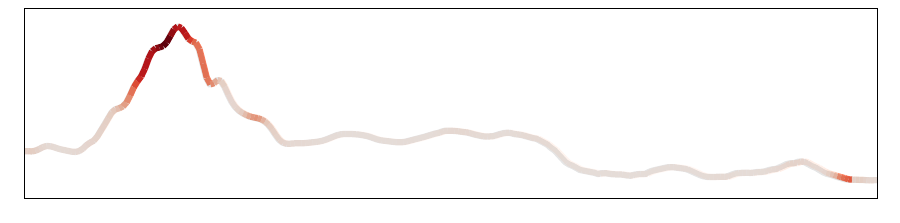}}
    \subfloat[acc: 0.9541]{\includegraphics[clip, trim=0cm 0cm 0cm 1.3cm, width=0.23\textwidth]{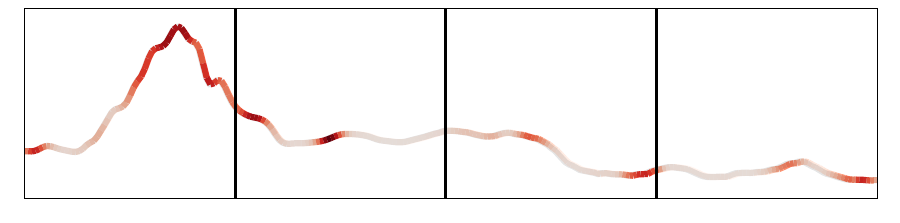}}
    \subfloat[acc: 0.9622]{\includegraphics[clip, trim=0cm 0cm 0cm 1.3cm, width=0.23\textwidth]{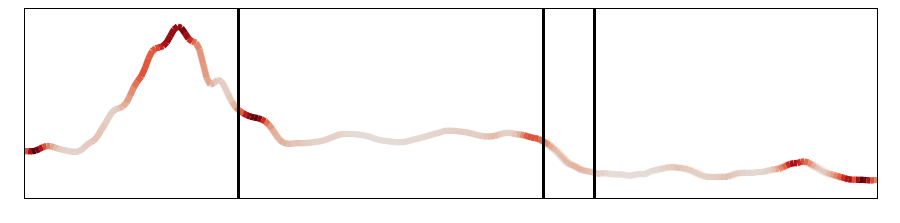}}
    \subfloat[acc: \bf 0.9676]{\includegraphics[clip, trim=0cm 0cm 0cm 1.3cm, width=0.23\textwidth]{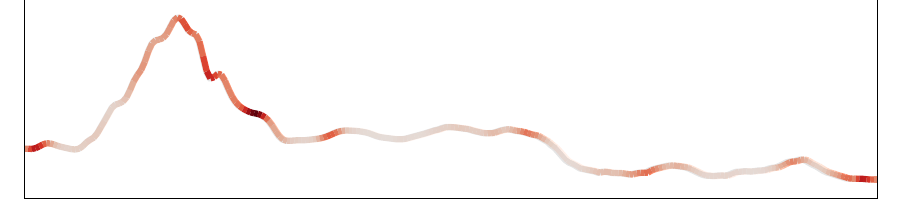}}\\

    \subfloat[acc: 0.9408]{\includegraphics[clip, trim=0cm 0cm 0cm 1.3cm, width=0.23\textwidth]{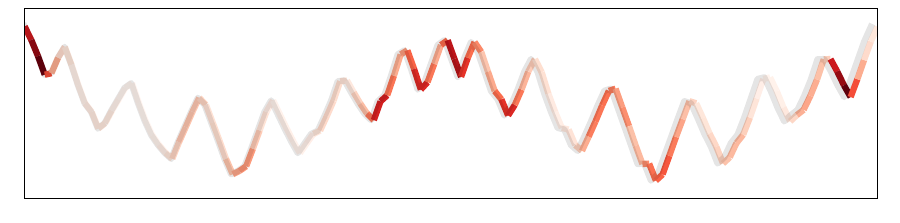}}
    \subfloat[acc: \bf 0.9552]{\includegraphics[clip, trim=0cm 0cm 0cm 1.3cm, width=0.23\textwidth]{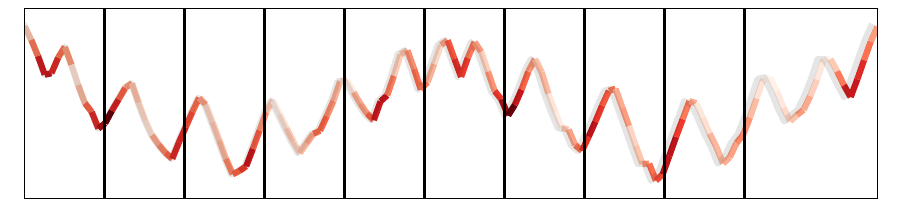}}
    \subfloat[acc: 0.9376]{\includegraphics[clip, trim=0cm 0cm 0cm 1.3cm, width=0.23\textwidth]{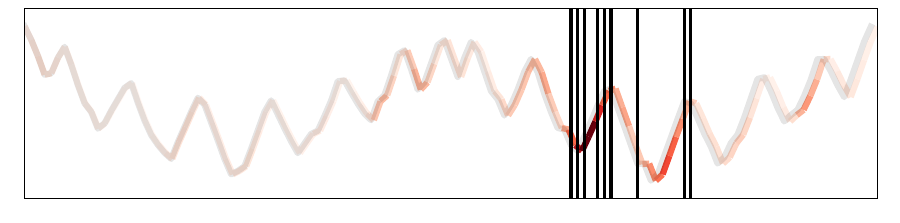}}
    \subfloat[acc: 0.9520]{\includegraphics[clip, trim=0cm 0cm 0cm 1.3cm, width=0.23\textwidth]{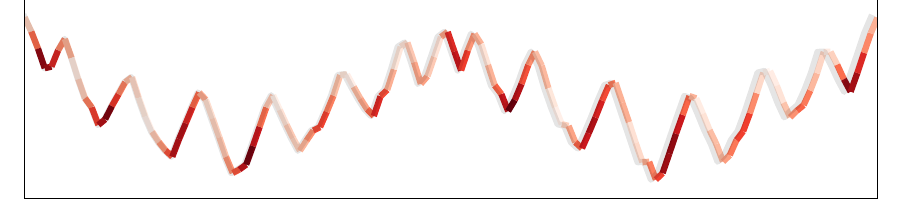}}\\

    \subfloat[acc: 0.7231]{\includegraphics[clip, trim=0cm 0cm 0cm 1.3cm, width=0.23\textwidth]{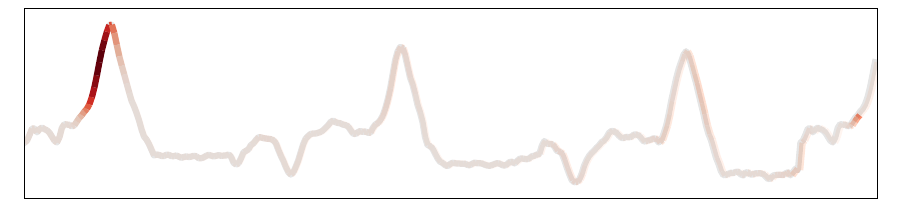}}
    \subfloat[acc: \bf  0.8769]{\includegraphics[clip, trim=0cm 0cm 0cm 1.3cm, width=0.23\textwidth]{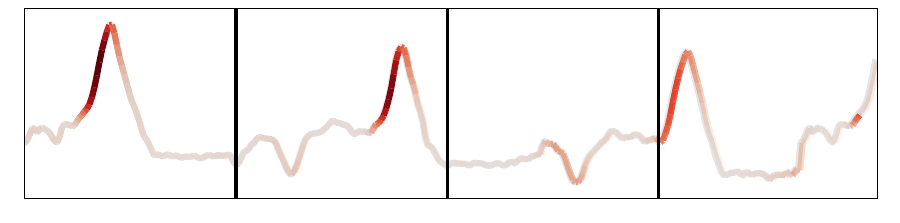}}
    \subfloat[acc: 0.7923]{\includegraphics[clip, trim=0cm 0cm 0cm 1.3cm, width=0.23\textwidth]{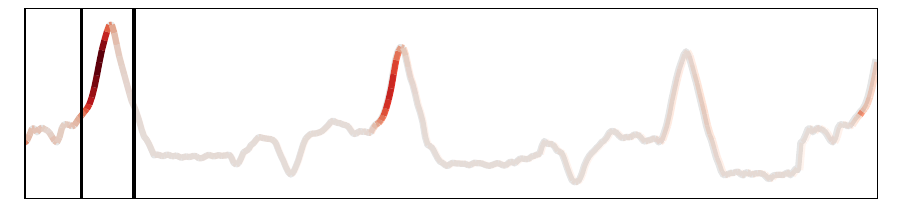}}
    \subfloat[acc: 0.8692]{\includegraphics[clip, trim=0cm 0cm 0cm 1.3cm, width=0.23\textwidth]{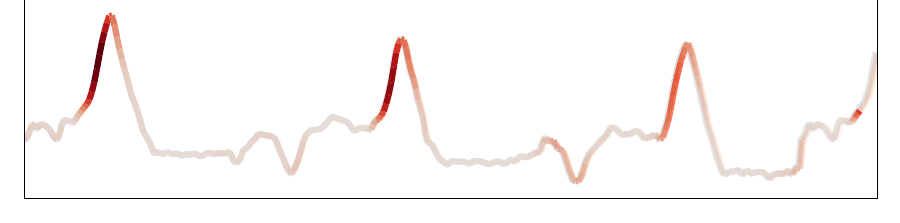}}\\

    \caption{Comparing LRP Input Attribution: Fixed vs Diverse Perspective Learning. This figure represents the input attribution (LRP) of distinct temporal poolings on the UCR repository, specifically for the Crop, Strawberry, SwedishLeaf, and ToeSegmentation2 datasets. Following the row sequence, these instances correspond to two samples of SoM-TP's best and STP's best scenarios, respectively. The figure highlights that SoM-TP excels when learning from diverse perspectives. Even when it doesn't rank as the best, the input attribution closely resembles that of the best independent pooling, ensuring robust performance.}
    \label{fig:extend4}
\end{figure*}

\clearpage
\subsection{Extended Experiment on Real-World Large Datasets}
\label{appendix1.3}
\subsubsection{Real-world Large Datasets in Time Series Classification: EEG and ECG}

\begin{figure*}[h]
\centering
    \subfloat[ECG]{\includegraphics[width=0.43\textwidth]{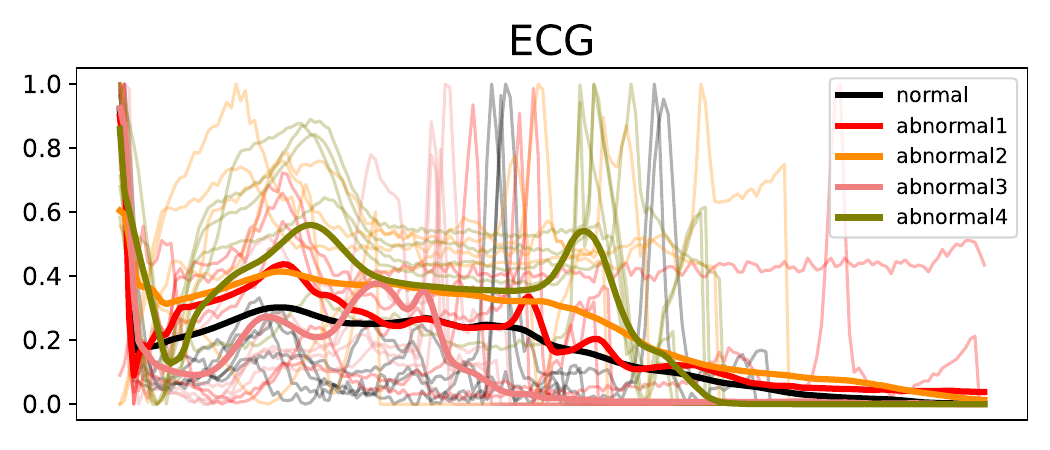}}
    \subfloat[EEG]{\includegraphics[width=0.43\textwidth]{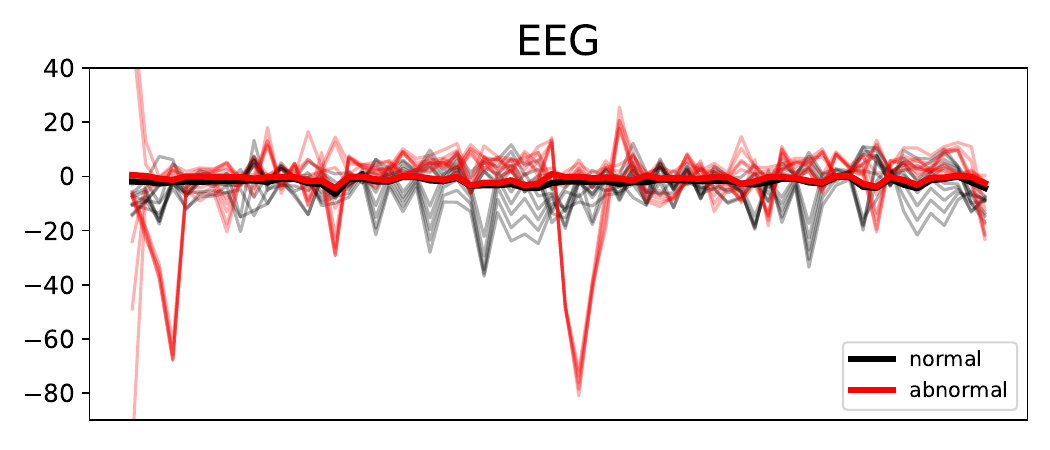}}
    \caption{ECG and EEG Datasets. First, ECG is univariate time series data with 5 classes, as shown in (a), with the x-axis representing time and the y-axis representing values. Second, EEG is multivariate time series data with 2 classes, non-alcoholic vs. alcoholic, as shown in (b), with the x-axis representing 65 channels and the y-axis representing values.}
\end{figure*}

\begin{table*}[h]
    \centering
    \begin{adjustbox}{width=0.9\textwidth}
    \begin{tabular}{lllllllllllll}
        \noalign{\smallskip}\noalign{\smallskip}\hline
        \bf Dataset & \bf Type & \bf \#Size & \bf \#Class & \bf \#Conv. & \multicolumn{2}{l}{\bf \#Pooled Prototypes} & \bf \#FC. & \bf Optim. & \bf lr & \bf batch & \bf window & \bf epoch\\
        &&&&& GTP & STP\&DTP &&&&&&\\
        \hline
        ECG & uni-variate & 109,446 & 2 & 1 & \multirow{2}{*}{1} & \multirow{2}{*}{$\mathbf{n}$} & \multirow{2}{*}{1} & \multirow{2}{*}{Adam} & \multirow{2}{*}{$1e-4$} & \multirow{2}{*}{$256$} & $1$ & \multirow{2}{*}{$100$}\\ 
        EEG & multi-variate (65) & 122,880 & 5 & 3 &&&&&&& $10$ & \\ 
        \hline
    \end{tabular}
    \end{adjustbox}
    \caption{Detailed data description and experimental settings.}

    \bigskip
    \begin{adjustbox}{width=0.9\textwidth}
    \begin{tabular}{ll|llll|llll}
        \noalign{\smallskip}\noalign{\smallskip}
        \multicolumn{2}{l}{\bf POOL (type)} & \multicolumn{4}{l}{\bf ECG (uni-variate)} & \multicolumn{4}{l}{\bf EEG (multi-variate)} \\
        & & ACC & F1 macro & ROC AUC & PR AUC & ACC & F1 macro & ROC AUC & PR AUC\\
        \hline
        \multirow{2}{*}{GTP} & MAX & 0.9236 & 0.7384 & 0.9854 & 0.9034 & \underline{0.9323} & \underline{0.9321} & \underline{0.9727} & \underline{0.9734} \\
                 & AVG & 0.9371 & 0.7839 & 0.9914 & \underline{0.9347} & 0.8821 & 0.8818 & 0.9426 & 0.9435 \\
                 \multirow{2}{*}{STP} & MAX & 0.9421 & 0.7866 & \underline{0.9918} & 0.9392 & 0.9162 & 0.9160 & 0.9663 & 0.9645 \\
                 & AVG & 0.9340 & 0.7611 & 0.9908 & 0.9222 & 0.8866 & 0.8865 & 0.9533 & 0.9514 \\
                 \multirow{4}{*}{DTP} & MAX-euc & 0.8593 & 0.6991 & 0.9856 & 0.9022 & 0.9087 & 0.9084 & 0.9619 & 0.9630 \\
                 & AVG-euc & \underline{0.9508} & \underline{0.8000} & 0.9860 & 0.9063 & 0.9108 & 0.9105 & 0.9628 & 0.9628 \\
                 & MAX-cos & 0.9179 & 0.7308 & 0.9841 & 0.9027 & 0.9212 & 0.9209 & 0.9694 & 0.9663 \\
                 & AVG-cos & 0.8802 & 0.6979 & 0.9849 & 0.8883 & 0.8747 & 0.8736 & 0.9362 & 0.9389 \\
                \hline
                 \multirow{2}{*}{SoM-TP} & MAX & \bf 0.9566 & \bf 0.8266 & \bf 0.9923 & \bf 0.9413 & \bf 0.9510 & \bf 0.9508 & \bf 0.9829 & \bf 0.9817\\
                                       & AVG & 0.9272 & 0.7765 & 0.9884 & 0.9129 & \bf 0.9424 & \bf 0.9421 & \bf 0.9799 & \bf 0.9810 \\
        \hline
    \end{tabular}
    \end{adjustbox}
    \caption{Model Performances. The best performance of SoM-TP is bolded, while the best performances of other temporal poolings are underlined.}
    \label{t3}
\end{table*} 

The UCR/UEA repository contains an extensive collection of TSC datasets. However, the repository's largest dataset size reaches only 30,000, which is relatively small. Thus, we further experiment on a large real-world dataset. Specifically, we choose the ECG dataset \cite{DBLP:journals/corr/abs-1805-00794} and the EEG dataset \cite{bay2000uci}, both of which are highly representative in the medical field. ECG pertains to univariate and multi-class classification, while EEG involves multivariate and binary classification tasks. These datasets comprise more than 100,000 data points each, and in the case of the multivariate dataset, the number of feature columns is 65. Therefore, this experiment encompasses datasets that are not covered by the existing benchmark repository.

\subsubsection{Experimental Settings}
First, for the ECG dataset, optimal results are achieved by adopting a shallow model with only one convolutional layer, as opposed to a deep model with multiple layers. Conversely, the EEG dataset demonstrates superior performance with FCN utilizing three convolutional layers. Regarding the batch size, a uniform setting of 256 is applied, taking into consideration the dataset size. For the EEG dataset, a window size of 10 is employed, meaning the previous 10 time steps serve as input for the model.

\subsubsection{Performance Analysis}
SoM-TP MAX demonstrates outstanding performance across both the EEG and ECG datasets. Notably, we extend the evaluation beyond accuracy, utilizing F1-score, ROC-AUC, and PR-AUC to address class imbalance issues. Remarkably, SoM-TP consistently outperforms existing temporal pooling methods in all of these evaluation metrics. Moreover, the experiment reveals that DTP is not always dominant over GTP or STP. Nevertheless, SoM-TP exhibits robust performance regardless of dataset conditions.

\clearpage
\twocolumn
\section{DTP Algorithm}
\label{app_dtp}
DTP is a pooling layer optimized by soft-DTW for dynamic segmentation considering the temporal relationship. The DTW distance is calculated through point-to-point matching with temporal consistency,
\begin{equation}
\begin{split}
&\mathbf{DTW}_\gamma(X, Y) = \mathbf{min}_\gamma\{ \langle A, \Delta(X, Y) \rangle, \forall A \in \mathcal{A} \}, \\
&\mathbf{min}_\gamma \{ a_1, ..., a_n \} =
\begin{cases}
\text{min}_{i \leq n} a_i, & \gamma=0 \\
-\gamma \text{log} \sum_{i=1}^{n} e^{-a_i / \gamma}, & \gamma > 0,
\end{cases}
\end{split}
\end{equation}
where $\mathbf{X}$ and $\mathbf{Y}$ are time series with lengths $\mathbf{t_1}$ and $\mathbf{t_2}$, and the cost matrix $\Delta (\mathbf{X, Y}) \in \mathbb{R}^{\mathbf{t_1} \times \mathbf{t_2}}$ represents the distance between $\mathbf{X_{t_1}}$ and $\mathbf{Y_{t_2}}$. $\mathbf{DTW}$ is defined as the minimum inner product of the cost matrix with any binary alignment matrix $\mathcal{A} \in {0, 1 }^{\mathbf{t_1} \times \mathbf{t_2}}$~\cite{lee2021learnable,cuturi2017soft}. With the soft-DTW algorithm, similar time sequences are grouped with different lengths.

\newpage
\begin{algorithm}[H]
\caption{Dynamic Temporal Pooling}         
\DontPrintSemicolon

  \SetKwFunction{FMain}{DTP}
  \SetKwFunction{forward}{forward}
  \SetKwFunction{backward}{backward}
  \SetKwFunction{optimize}{optimization}
  \SetKwFunction{optim}{optimize}
 
  \SetKwProg{Fn}{Function}{:}{}
  \Fn{\FMain{$\mathbf{P}$, $\mathbf{H}$}}{
  
        $\delta(p_l, h_t) = 1 - \frac{p_l \cdot h_t}{||p_l||_2 ||h_t||_2}$
        
        \Fn{\forward{$\mathbf{P}$, $\mathbf{H}$}}{
            $\triangleright$ fill the alignment cost matrix $R \in \mathbb{R}^{L\times T}$
            
            $R_{0,0}=0, R_{:,0}=R_{0,:}=\infty$
            
            \For{$l=1, ..., L$}{
                
                \For{$t=1, ..., T$}{
                    $R_{l, t} = \delta(p_l, h_t) + min_\gamma\{R_{l-1,t-1}, R_{l, t-1}\}$
                }
            }
            \textbf{return} $DTW_\gamma(\mathbf{P, H}) = R_{L, T}$
        }
        
        \;
        
        \Fn{\backward{$\mathbf{P}$, $\mathbf{H}$}}{
            $\triangleright$ fill the soft alignment matrix $E \in \mathbb{R}^{L\times T}$
            
            $E_{l, t}=\partial R_{L, T}/\partial R_{l,t}$
            
            $E{:, T+1} = E_{L+1, :} = 0$
            
            $R_{:, T+1} = R_{L+1,:} = -\infty$
            
            \For{$l = L, ..., 1$}{
                \For{$t = T, ..., 1$}{
                    $a = exp\frac{1}{\gamma}(R_{l, t+1} - R_{l, t} - \delta(p_l, h_{t+1}))$ \;
                    
                    $b = exp\frac{1}{\gamma}(R_{l+1, t+1} - R_{l, t} - \delta(p_{l+1}, h_{t+1}))$ \;
                    
                    $E_{l,t} = a \cdot E_{l, t+1} + b \cdot E_{l+1, t+1}$
                }
            }
            
            \textbf{return} $\nabla_P DTW_\gamma(\mathbf{P, H}) = (\frac{\partial \Delta(\mathbf{P, H})}{\partial \mathbf{P}})^T E$
        }
        
        \Fn{\optimize{$(\mathbf{X}, y), \mathbf{P}, \Phi$}}{
            $\triangleright \mathcal{W}$: network $\Phi$ parameter
            
            $\triangleright w^{(c)} \in \mathbb{R}^K$  of class weight vector
            
            $\triangleright \mathbf{W}^{c} = [w_1^{(c)}, ..., w_L^{(c)}] \in \mathbb{R}^{k\times L}$ of class weight matrix
            
            $\triangleright P(y=c|\mathbf{X}) = \frac{exp(\sum_{l=1}^{L} \bar{h_l} \cdot w_l^{(c)})}{\sum_{c'=1}^L exp(\sum_{l=1}^L \bar{h_l} \cdot w_l^{c'})}$ of posterior

            $\mathcal{L}_{proto}(\mathbf{P}) = \frac{1}{N} \sum_N DTW_\gamma (P, \Phi(\mathbf{X}^n ; \mathcal{W}))$
            
            $\mathcal{L}_{class}(\mathcal{W}, \{\mathbf{W}^{(c)}\}) \\ = -\frac{1}{N}\sum_{n=1}^N log P(y=y^n|\mathbf{X}^n)$
            
            \;
            
            \Fn{\optim{$\mathbf{P, W}, \mathcal{W}$}}{
                $\mathbf{P} \leftarrow \mathbf{P} - \eta \cdot \partial \mathcal{L}_{proto}/\partial \mathbf{P}$
                
                $\mathcal{W} \leftarrow \mathcal{W} - \eta \cdot \mathcal{L}_{class} / \partial \mathcal{W}$
                
                $\mathbf{W}^{(c)} \leftarrow \mathbf{W}^{(c)} - \eta \cdot \partial \mathcal{L}_{class} / \partial \mathbf{W}^{(c)}$}
        }
    }
\label{algo}  
\end{algorithm}

\end{document}